\documentclass[accepted]{uai2025} 

\usepackage{natbib}
\usepackage{mathtools}
\usepackage{booktabs}
\usepackage{tikz}
\usepackage{multirow}
\usepackage{amsthm}
\usepackage{algpseudocode}
\usepackage{comment}
\usepackage{adjustbox}
\usepackage{arydshln}
\usepackage{xcolor}
\usepackage{amsmath}
\usepackage{algorithm}
\usepackage{algorithmicx}
\usepackage{subcaption}
\usepackage{amsfonts}

\newtheorem{proposition}{Proposition}
\newtheorem{lemma}{Lemma}
\newtheorem{corollary}{Corollary}

\DeclareMathOperator*{\argmax}{arg\,max}
\DeclareMathOperator*{\argmin}{arg\,min}
\newcommand{\xnotS}{\mathbf{x}_{\bar{S}}}
\newcommand{\xS}{\mathbf{x}_{S}}

\title{Model-Agnostic Dynamic Feature Selection with Uncertainty Quantification}

\author[1]{Javier~Fumanal-Idocin}
\author[2]{Raquel~Fernandez-Peralta}
\author[1]{Javier~Andreu-Perez}

\affil[1]{%
    School of Computer Science and Electronic Engineering\\
    University of Essex\\
    Colchester, Essex CO4 3SQ, United Kingdom\\
    \texttt{\{j.fumanal-idocin, j.andreu-perez\}@essex.ac.uk}
}
\affil[2]{%
    Institute of Mathematics\\
    Slovak Academy of Sciences\\
    Štefánikova 49, 814 73 Bratislava, Slovakia\\
    \texttt{raquel.fernandez@mat.savba.sk}
}

\begin{document}
\maketitle

\begin{abstract}
Dynamic feature selection (DFS) addresses budget constraints in decision-making by sequentially acquiring features for each instance, making it appealing for resource-limited scenarios. However, existing DFS methods require models specifically designed for the sequential acquisition setting, limiting compatibility with models already deployed in practice. Furthermore, they provide limited uncertainty quantification, undermining trust in high-stakes decisions. In this work, we show that DFS introduces new uncertainty sources compared to the static setting. We formalise how model adaptation to feature subsets induces epistemic uncertainty, how standard imputation strategies bias aleatoric uncertainty estimation, and why predictive confidence fails to discriminate between good and bad selection policies. 
We also propose a model-agnostic DFS framework compatible with pre-trained classifiers, including interpretable-by-design models, through efficient subset reparametrization strategies. Empirical evaluation on tabular and image datasets demonstrates competitive accuracy against state-of-the-art greedy and reinforcement learning-based DFS methods with both neural and rule-based classifiers. We further show that the identified uncertainty sources persist across most existing approaches, highlighting the need for uncertainty-aware DFS.
\end{abstract}


\section{Introduction}

Many real-world decision-making scenarios involve prohibitively high-dimensional information spaces, where acquiring all possible data is unfeasible due to cost or time constraints \citep{erion2022cost}. Consider, for instance, a doctor diagnosing a patient. The doctor must select only the right clinical tests to conduct, as they are expensive in terms of both money and time, which are usually scarce resources. Only a small subset will be truly relevant in accurately diagnosing the patient's specific condition, and this subset will be different for each patient. The most effective approach in such situations is to adapt the information acquisition process to the subject's condition. This methodology, where features are acquired dynamically based on their evolving relevance, is commonly referred to as Dynamic Feature Selection (DFS) \citep{melville2004active}.

The existing literature on DFS primarily explores two main approaches: reinforcement learning (RL) \citep{janisch2019classification} and greedy optimisation \citep{chen2015sequential}. RL-based approaches offer the potential to discover optimal policies for feature acquisition, but they come with the inherent challenges of RL training, including significant computational demands and potential convergence issues \citep{erion2022cost}. Greedy optimisation methods maximise a measure that relates the known inputs with the response variable \citep{chattopadhyayvariational, takahashi2025dynamic}. Usually, this measure is the conditional mutual information (CMI) between input variables and the response variable, which has some desirable qualities in terms of uncertainty minimisation and prediction power \citep{gadgilestimating}. These approaches can focus on the CMI-estimation task directly \citep{gadgilestimating}, or by means of generative models \citep{ma2018eddi}, also circumvent the issue of selecting a discrete variable in a neural network, which has non-differentiability issues \citep{covert2023learning}. Nevertheless, both approaches rely on predictors that operate across a large number of feature subsets. This requirement is closely related to the core problem studied in Invariant Risk Minimisation \citep{krueger2021}, where a single model must generalise across multiple environments. These subset‐adaptive methods allow specialisation while retaining shared knowledge across subsets. Examples of this approach are hierarchical or modular parametrisations \citep{rebuffi2018}, mixture-of-experts architectures \citep{shazeer2017}, or meta-learning strategies \citep{finn2017}. 

Despite the benefits of DFS in budget-sensitive scenarios, existing methods have struggled with two major limitations that hinder their application in real-life scenarios:
\begin{itemize}
    \item Incomplete Uncertainty Quantification: existing DFS methods optimise feature selection for predictive performance, but they do not quantify the epistemic uncertainty in their decisions, and they do not validate the calibration of their probability estimates during sequential acquisition. This gap prevents practitioners from distinguishing between predictions that are confidently correct versus those that merely appear plausible.
    \item Incompatibility with Pre-trained Models: Existing DFS methods require classifiers specifically designed to handle arbitrary feature subsets, making them incompatible with models already validated and deployed in practice. For example, interpretable-by-design classifiers, whose transparency is critical in high-stakes domains such as clinical diagnosis. Adapting pre-trained models to the sequential acquisition setting can be computationally expensive or architecturally infeasible, creating a significant barrier between DFS research and real-world deployment.
\end{itemize}

In this paper, we address both limitations with the following contributions:
\begin{itemize}
    \item We formally analyse the sources of uncertainty introduced by dynamic feature selection. We characterise subset-dependent epistemic uncertainty arising from model adaptation, identify biases in aleatoric uncertainty estimation during sequential acquisition, and show why predictive entropy cannot reliably discriminate between good and bad selection policies.
    
    \item We propose a model-agnostic DFS framework that explicitly incorporates epistemic uncertainty into the feature acquisition process, enabling efficient subset adaptation of pre-trained and interpretable-by-design classifiers with minimal computational overhead.
    
    \item We empirically evaluate our framework on five tabular and three image datasets, demonstrating competitive accuracy against state-of-the-art greedy and reinforcement learning–based DFS methods. We further show that the identified uncertainty sources persist across existing DFS approaches.
\end{itemize}

\section{Uncertainty Quantification in Dynamic Feature Selection} \label{sec:uncertainty_in_dfs}

\paragraph{Notation and Problem Formulation.} Let $\mathbf{x} = (x_1, \dots, x_M)$ be an input feature vector and $y \in \{1, \dots,C\}$ a corresponding target label  in a supervised learning paradigm. The complete set of $N$ input samples is denoted as $X = \{\mathbf{x}^1, \dots, \mathbf{x}^N\}$. A partially observed sample, where only features indexed by $S$ have been observed, is denoted as $\mathbf{x}_S = \{x_i: i \in S\}$. The set of unobserved features indices is denoted as $\bar{S} = \{1,\dots,M\} \setminus S$.

\paragraph{Uncertainty Quantification in Machine Learning.} Following the taxonomy given in \cite{hullermeier2021aleatoric}, uncertainty can be categorized into two main categories. Aleatoric Uncertainty is the inherent irreducible noise in the data-generating process. This may arise from the analysed process itself, i.e., tossing a coin, noise in the measurements, etc. Epistemic Uncertainty is the model's uncertainty due to a lack of knowledge about the optimal parameters $\theta$, which can be reduced with more data. Total uncertainty is the addition of both, and it is usually expressed as the entropy of the prediction of the model.

DFS introduces complications to this standard taxonomy. Unlike static feature selection, where uncertainty is assessed once on a fully observed input, DFS requires uncertainty quantification at each step of a sequential decision process, with predictions conditioned on dynamically acquired feature subsets $\mathbf{x}_S$. As a result, both aleatoric and epistemic uncertainty become subset-dependent and evolve throughout the acquisition process. Moreover, adapting a single predictive model to an exponential number of feature subsets introduces additional sources of epistemic uncertainty that are not present in the static setting. In the following subsections, we formalise these effects and analyse how they impact prediction reliability and feature acquisition in DFS.

\subsection{Model Adaptation creates Epistemic Uncertainty} \label{sec:model_adaptation}


In the DFS literature, the classifier parameters, $\theta$, are usually trained  using the following expression \citep{gadgilestimating}:
\begin{equation} \label{eq:basic_dfs_loss2}
    \theta^*  = \arg \min_{\theta}  \mathbb{E}_ S \mathbb{E}_{\mathbf{x}, y}  l(f_\theta(\mathbf{x}_S), y),    
\end{equation}
where $f_{\theta}$ denotes a model capable of making predictions given any subset of features $S$. To evaluate $f_\theta(\mathbf{x}_S)$, missing features are typically handled through some form of imputation method, such as zero/mean or conditional expectation imputation. Although solving Eq. (\ref{eq:basic_dfs_loss2}) results in a parameter configuration that minimises the expected loss across all possible subsets, the resulting model may differ from a local model trained exclusively on a particular subset of features with an equivalent training objective:
\begin{equation}
    \theta_S^* = \arg \min_{\theta_S}   \mathbb{E}_{\mathbf{x}, y}  l(f_{\theta_S}(\mathbf{x}_S), y).
\end{equation}
If we measure the risk of a model on subset $S$ as $\mathcal{R}_S(f) = \mathbb{E}_{\mathbf{x}, y} l(f(\mathbf{x}_S), y)$, we can expect both models to have different risks. We measure the discrepancy of $f_{\theta^*}$ and $f_{\theta_S^*}$ as their difference in risk:
\begin{equation} \label{eq:subset_risk_gap}
\Delta_S = \mathcal{R}_S(f_{\theta^*}) - \mathcal{R}_S(f_{\theta_S^*}).
 \end{equation}
A positive $\Delta_S$ indicates that the globally trained model $f_{\theta^*}$ performs worse on the specific subset $S$ than a hypothetical local optimum $f_{\theta_S^*}$, reflecting a loss of specialisation caused by averaging across subsets in Eq. \eqref{eq:basic_dfs_loss2}. Conversely, $\Delta_S \approx 0$ implies that $f_{\theta^*}$ generalizes well to $S$, achieving near‐optimal performance despite joint training. 

The exponential growth of the feature subset space, $2^M$, forces $\theta^*$ to balance many local optima and unseen feature subsets, which may lead to large $\Delta_S$. Because of this, it is common in the literature to, instead of sampling uniformly $S$, train jointly the predictor with the feature selector  $q_{\phi}$. This changes Eq. (\ref{eq:basic_dfs_loss2}) into:
\begin{equation} \label{eq:basic_dfs_loss_biased}
    \theta^*  = \arg \min_{\theta} \mathbb{E}_{\mathbf{x}, y} \mathbb{E}_{S \sim q_{\phi}(\cdot|\mathbf{x})}   l(f_\theta(\mathbf{x}_S), y).
\end{equation}
However, this approach also has problems. The selector $q_{\phi}$ may explore only a small fraction of all possible feature subsets $S$. This results in a trade-off where predictive power is diminished for rarely selected subsets. Moreover, this scheme is only effective when the predictor guides the feature selection process, but it is vulnerable to changes in external factors. For example, when relevant variables needed by the selector are unavailable for certain samples.


\subsection{Estimating Aleatoric Uncertainty during Sequential Acquisition} \label{sec:aleatoric_estimation}

The ideal DFS classifier should properly estimate $p(y|\mathbf{x}_S)$ for every $S$ in the query process: 

\begin{equation} \label{eq:p_s_estimation}
    p(y \mid \mathbf{x}_{S}) = \int_{\mathbf{x}_{\bar{S}}} p(y \mid \mathbf{x}_{S}, \mathbf{x}_{\bar{S}}) p(\mathbf{x}_{\bar{S}} \mid \mathbf{x}_{S}) \, d\mathbf{x}_{\bar{S}}.
\end{equation}

However, many classical imputation methods for the unobserved features bias this estimation. For example, if we use the classical mean imputation method, we are implicitly assuming a degenerate distribution where the unobserved features are fixed at their expected values, $\mu(\mathbf{x}_{\bar{S}})$. Consequently, Eq. (\ref{eq:p_s_estimation}) simplifies to $p(y \mid \mathbf{x}_{S}, \mathbf{x}_{\bar{S}} = \mu(\mathbf{x}_{\bar{S}}))$, which biases the prediction.

In previous DFS literature, there have been two options to solve this issue. One is to use a generative model to estimate $p(\xnotS \mid \xS)$ \citep{ma2018eddi}. However, this approach requires a lot of data to work properly and tends to overfit. The alternative approach consists of imputing the missing values in $\mathbf{x}_{\bar{S}}$ with a fixed value. Then, the input to the classifier is modified by concatenating the observed feature vector with a mask that indicates which features are really observed. This technique allows the model to differentiate between imputed and truly observed features. Nonetheless, the effectiveness of augmenting the input with a missingness mask depends on how informative the missingness pattern is with respect to the response. When missingness carries little additional signal, the mask variables contribute minimally to the prediction, and the resulting model behaves similarly to standard mean imputation \cite{VanNess2023}.

\subsection{Non-Monotonicity of Uncertainty in Feature Acquisition} \label{sec:non_monotinicy}
A common implicit assumption in dynamic feature selection is that acquiring additional features monotonically improves prediction quality and reduces uncertainty. However, neither prediction uncertainty nor model risk is guaranteed to evolve monotonically. For instance, in the diagnosis of a rare disease ($0.1\%$ prevalence), the prior predictive distribution is highly skewed and low-entropy, however, epistemic uncertainty is also very high. After observing a relevant but uncertain test result, the posterior may approach a balanced distribution, increasing predictive entropy and reducing confidence, even though the epistemic uncertainty was reduced.

Epistemic uncertainty is also not guaranteed to decrease when querying additional features, due to model adaptation effects. While it is commonly assumed that predictions made with $S \cup \{i\}$ are more reliable than those made with $S$, this requires the monotonicity condition $\mathcal{R}_{S \cup \{i\}}(f_{\theta^*}) \leq \mathcal{R}_S(f_{\theta^*})$ to hold. However, this condition is highly problem-dependent and might be violated in practice, particularly when a single model must adapt to many different feature subsets. Existing DFS methods rely on jointly training feature selectors and predictors to mitigate this effect \citep{chattopadhyayvariational}. However, distribution shift and limited exploration in training can reintroduce the non-monotonic behaviour of model risk.

\section{A Model-Agnostic Dynamic Feature Selection Procedure}

Directly training models for DFS is not always practical: large pre-trained models are expensive to fine-tune~\cite{kumar2022fine}, and subset-specific adaptation of interpretable models such as rule-based classifiers incurs prohibitive overhead (Appendix~\ref{apx:dfs_rules_adaption_naive}). Besides, practitioners often need to incorporate models that have already been validated in deployment, where retraining is neither feasible nor desirable.

We therefore propose a model-agnostic DFS framework built around a single global predictor $f_\theta$. When the model is not inherently compatible with partial observations, we introduce an adaptation mechanism based on input imputation and subset-dependent reparametrization, enabling the fixed predictor to operate over arbitrary feature subsets without full retraining. Our DFS policy also explicitly estimates the epistemic uncertainty induced by working with partial information, which allows the DFS to select those features where decisions are most reliable.

\subsection{Model Reparametrization for Feature Subsets} \label{sec:reparametrization}

The challenges identified in Section~\ref{sec:uncertainty_in_dfs} become particularly relevant when working with a pre-specified global model $f_\theta$ that was not designed for dynamic feature selection. Such models expect inputs in the full feature space and therefore cannot natively process arbitrary observed subsets $\mathbf{x}_S$.

To enable evaluation on any subset $S$, we complete the partially observed input by imputing the missing features and then evaluate the original model on the imputed representation. This allows the fixed model $f_\theta$ to produce predictions for arbitrary subsets. However, these predictions depend on the imputation mechanism and generally differ from the true conditional $p(y \mid \mathbf{x}_S)$, introducing an approximation bias. To reduce this discrepancy, we introduce a subset-dependent reparametrization of the global model:
\begin{equation}
\theta_S = R_\psi(\theta, S),
\end{equation}
where $R_\psi$ is a reparametrization function with parameters $\psi$, trained to minimize the expected risk across feature subsets:
\begin{equation}
\psi^* = \arg\min_\psi 
\mathbb{E}_S \, \mathbb{E}_{x,y} 
l\big(f_{R_\psi(\theta,S)}(\mathbf{x}_S), y\big).
\end{equation}
This reparametrization reduces the subset-specific risk gap $\Delta_S$ and mitigates the bias induced by imputation, bringing predictions closer to the true conditional $p(y \mid \mathbf{x}_S)$. Compared to full subset-specific training, this approach is computationally efficient, preserves the original parameters of $f_\theta$, and avoids degrading a previously validated model. For neural networks, $R_\psi$ can adapt only the final layer or implement low-rank updates~\citep{hu2022lora}. Rule-based classifiers admit particularly efficient reparametrization strategies, detailed in Appendix~\ref{appendix:A}.

\subsection{Defining the DFS policy given a global model} \label{sec:rules_dfs_policy}

Our DFS policy aims to sequentially select features so as to approximate the global model's predictions for each sample. For that, we define a value function, $v_q(i, \mathbf{x}_S)$, that computes the expected added value of adding the feature $i$ given the currently observed $\mathbf{x}_S$. At each step of the DFS, we compute this expression for each unobserved feature.

To compute $v_q(i, \mathbf{x}_S)$ we use a function $q(x_i,\mathbf{x}_S)$ that quantifies the improvement of the prediction quality of $f_{\theta, S\cup \{i\}}$ with respect to the prediction with all features, $f_{\theta}$:
\begin{equation} \label{eq:main_value_function}
	v_q(i, \mathbf{x}_S) = \mathbb{E}_{p(x_i \mid \mathbf{x}_S)} \left[ q(x_i, \mathbf{x}_{S})\right].    
\end{equation}
Function $q(x_i, \mathbf{x}_S)$ combines two components: the predictive improvement from adding feature $i$, captured by $u(\mathbf{x}_S \cup \{x_i\})$, and the epistemic uncertainty of the resulting model, captured by $e(\mathbf{x}_S \cup \{x_i\})$:
\begin{equation} \label{eq:gain}
	q(x_i, \mathbf{x}_S) = u(\mathbf{x}_S \cup \{x_i\}) - \lambda e(\mathbf{x}_S \cup \{x_i\}).  
\end{equation}
\noindent The parameter $\lambda$ regulates the relative importance of epistemic uncertainty in the DFS process.

We can opt to have a greedy or an RL approach. If we use a greedy policy, we just choose the feature that maximises $v_q$. To use RL we formulate the problem as a finite-horizon Markov decision process, where the state at time $t$ is the set of observed features $\mathbf{x}_{S_t}$, the action corresponds to selecting the next feature $i \in \overline{S_t}$ to query, and the immediate reward is given by $v_q(i,\mathbf{x}_{S_t})$. In this setting, the optimal action-value function $Q^*(\mathbf{x}_{S}, i)$ satisfies the Bellman optimality equation \citep{sutton1998reinforcement}:
\begin{align}\label{eq:main_value_function}
Q^*(\mathbf{x}_{S_t}, i)
&= v_q(i, \mathbf{x}_{S_t})
 \notag \\
&  ~~~~ + \gamma \,
\mathbb{E}_{p(x_i \mid \mathbf{x}_{S_t})}\Big[
\max_{i' \in \overline{S_t \cup \{i\}}}
Q^*(\mathbf{x}_{S_t \cup \{i\}}, i')
\Big]
\end{align}
where $\gamma \in [0,1]$ is the discount factor, and the expectation accounts for uncertainty over the value of the queried feature $x_i$. The policy then selects features according to $\pi^*(\mathbf{x}_S) = \arg\max_{i \in \overline{S}} Q^*(\mathbf{x}_S, i)$.

In both approaches, to support variable feature costs, we divide $v_q(x_i, \mathbf{x}_S)$ by the cost of the feature, so that the policy optimises $q$ per unit cost. This is not an optimal solution \cite{chen2015sequential}, but it works well in practice, and it is cheap to compute.

The next step is computing $v_q$. For that, we need to know: (i) how to compute $u$ using the global prediction (Section \ref{sec:subsec1}); (ii) how to estimate the epistemic uncertainty, $e$, for a given decision (Section \ref{sec:subsec2}); (iii) and how to estimate the $v_q(i, \mathbf{x}_S)$ value in test time (Section \ref{sec:subsec3}).

\subsubsection{Difference Between Global and sub-Model Prediction} \label{sec:subsec1}
For computing the prediction difference of the sub-model defined by $S$ with respect to the full model, we consider the Kullback–Leibler divergence of the model predictions:
\begin{equation} \label{eq:prediction_uncertainity}
    u(\mathbf{x}_S) = D_{\text{KL}}(p(\hat{y}|\mathbf{x}) \| p(\hat{y}|\mathbf{x}_S)).
\end{equation}
When an additional feature $x_i$ is queried, this becomes:
\begin{equation} \label{eq:prediction_uncertainityII}
    u(\mathbf{x}_S \cup \{x_i\}) = D_{\text{KL}}(p(\hat{y}|\mathbf{x}) \| p(\hat{y}|\mathbf{x}_S \cup \{x_i\})).
\end{equation}
\noindent As shown in Appendix~\ref{appendix:CMIcomparison}, the reduction in Eq.~\eqref{eq:prediction_uncertainity} induced by querying $x_i$ admits an information–theoretic interpretation and is related to conditional mutual information (CMI) \citep{covert2023learning, gadgilestimating, ma2018eddi}. Both quantities involve a log-ratio term comparing predictive distributions before and after observing $x_i$, but differ in the reference measure used for weighting. CMI quantifies the statistical dependence between $x_i$ and $\hat{y}$ conditioned on $\mathbf{x}_S$, whereas our criterion measures how much observing $x_i$ reduces the discrepancy between the subset-based predictor and a fixed global model. Although not equivalent in general, they align when $p(\hat{y}\mid \mathbf{x}_S,x_i)$ closely matches $p(\hat{y}\mid \mathbf{x})$, or when $x_i$ induces a highly concentrated posterior.

\subsubsection{Epistemic uncertainty estimation} \label{sec:subsec2}
Ensembles are considered state-of-the-art for epistemic uncertainty quantification, as prediction variance across members captures ignorance about a decision~\citep{rahaman2021uncertainty, ovadia2019can}. When feasible, this is the preferred solution. For large models where ensembles are computationally prohibitive, alternatives include MC-dropout~\citep{gal2016dropout}, provided the global model has dropout layers, or feature-space similarity metrics~\citep{liu2020simple}. For explainable models, ensembles are also not desirable, as aggregating predictions across ensemble members obscures the decision logic \citep{bassan2025makes}. In that case, we propose to keep the fixed primary predictor for decision-making and explanation, and use auxiliary models only to quantify epistemic uncertainty around that predictor’s output. Specifically, let $\mathcal{I}$ denote a set of auxiliary classifiers trained using different random initialisations or bootstrap samples. For a given sample with observed features $\mathbf{x}_S$, let $p_{\theta}(\hat{y}|\mathbf{x}_S)$ denote the prediction of the primary model with parameters $\theta$, and let $p_{\theta_i}(\hat{y}|\mathbf{x}_S)$ denote the prediction of auxiliary model $i$ with parameters $\theta_i$. We define epistemic uncertainty as the average divergence between the primary prediction and the auxiliary predictions:
\begin{equation} \label{eq:epistemic_normal}
    e(\mathbf{x}_S) = \frac{1}{|\mathcal{I}|} \sum_{i \in \mathcal{I}} D_{\text{KL}}\left( p_{\theta_i}(\hat{y} \mid \mathbf{x}_S \,\|\, p_{\theta}(\hat{y} \mid \mathbf{x}_S))\right).
\end{equation}
This expression quantifies how atypical the primary model’s predictive distribution is relative to a set of alternative plausible predictors. This makes it sensitive to situations where the primary model is confident yet unstable under training perturbations, as indicated by systematic disagreement with the auxiliary models. A further discussion on the adequacy of $e(\mathbf{x}_S)$ as a measure of epistemic uncertainty is provided in Appendix \ref{apx:metric_analysis}.

\subsubsection{Estimating the Value Function in Test Time} \label{sec:subsec3}
Since unqueried features are unavailable at inference, we train a neural estimator to approximate $v_q(i, \mathbf{x}_S)$ from observed features $\mathbf{x}_S$. The estimator uses a dual-head architecture to jointly predict both $u(\mathbf{x}_S \cup \{x_i\})$ and $e(\mathbf{x}_S \cup \{x_i\})$, trained by minimizing squared prediction error across all possible feature additions (architecture and training details in Appendix~\ref{app:neural_estimator}).

\section{Experimentation} \label{sec:experimentation}

\subsection{Datasets and methods}

Our empirical evaluation considers five tabular datasets from both medical and non-medical domains: \emph{Diabetes}, \emph{Heart Disease}, and \emph{Cirrhosis}; and \emph{Wine} and \emph{Yeast}. These datasets range from a few hundred to over $10^5$ samples and from 8 to 47 features, capturing substantial variability in both scale and dimensionality. For image datasets, we use CIFAR-10~\cite{krizhevsky2009learning}, which contains 60,000 32×32 colour images across 10 and 100 classes, respectively; MNIST~\cite{lecun2002gradient}, consisting of 70,000 28×28 grayscale images of 10 handwritten digits; and Imagenette~\cite{howard2019imagenette}, a subset of ImageNet with 13,394 images across 10 classes. Full dataset statistics for each tabular dataset are summarised in Appendix~\ref{appendix:report_datasets}. We measure predictive performance using classification accuracy and calibration quality using the Expected Calibration Error (ECE), which evaluates the alignment between predictive confidence and empirical accuracy across confidence bins.

We compare both static and dynamic feature selection approaches. For static selection, we include mutual information-based feature scoring with a decision tree model. We also include TabNet with budget constraints, which sees the full available feature and selects those relevant until the budget is exhausted \citep{arik2021tabnet}. For pure DFS, we evaluate: (i) CMI-based greedy optimisation (DIME) \citep{covert2023learning}, (ii) variational greedy information pursuit (VIP) \citep{chattopadhyayvariational}, (iii) Q-learning-based RL (CWCF) \citep{janisch2019classification}, and (iv) actor-critic networks (INVASE) \citep{yoon2018invase}. We evaluate our DFS adaptation method using different classifiers: a standard Multilayer Perceptron (MLP) with and without parametrisation as backbone, and two rule-based approaches: CART decision trees \citep{breiman2017classification} and fuzzy decision trees \citep{fumanal2025fast}. For tabular data, we use both greedy and RL-based DFS learnt policies. For image datasets, we only use greedy approaches for computational reasons. Hyperparameter selection and further details for all methods are provided in Appendix \ref{appendix:report_hyperparam}.

\subsection{Results for Tabular Data}
\begin{table*}[ht]
    \centering
    \caption{Accuracy comparison of Static and Dynamic Feature Selection methods for tabular datasets with variable feature costs. Results show the average accuracy across all budget levels (from 5\% to 50\% in evenly 5\% increments), with standard deviation indicating variability across budget levels. The final rank is computed as the mean of the ranks obtained across the five individual datasets. Methods marked with * are our proposed approaches.}
    \adjustbox{width=\linewidth}{
    \begin{tabular}{ccc|ccccc|c}
    \toprule
         &  & Method          &  Cirrhosis           &  Diabetes            &  Heart               &  Wine                &  Yeast               &  Average (rank)   \\
    \midrule
    \multirow{2}{*}{}  &
&  Static CMI            &  $72.62 \pm 1.68$ &  $62.63 \pm 0.13$ &  $65.58 \pm 9.27$ &  $79.17 \pm 1.97$ &  $31.65 \pm 3.52$          &  $62.33$ (6.0)    \\
 &              &  TabNet         &  $42.86 \pm 7.01$          &  $53.15 \pm 4.02$          &  $62.30 \pm 8.75$          &  $61.11\pm 10.67$          &  $31.65\pm 4.13$          &  $50.21$ (11.8)    \\
        \midrule
        \multirow{11}{*}{Dynamic} & \multirow{5}{*}{RL} &  ReMLP*         &  $70.12 \pm 1.72$ &  $60.35 \pm 1.81$ &  $75.08 \pm 9.80$ &  $90.28 \pm 3.98$ &  $37.24 \pm 4.65$ &  $66.61$ (2.8)    \\
                    & &  MLP*           &  $69.76 \pm 3.23$ &  $58.43 \pm 0.90$ &  $64.92 \pm 11.12$&  $86.39 \pm 3.06$ &  $39.02 \pm 6.44$ &  $63.71$ (5.2)    \\
                    & &  CWCF           &  $49.64 \pm 6.59$ &  $55.56 \pm 4.23$ &  $70.66 \pm 6.62$ &  $82.50 \pm 13.74$&  $32.52 \pm 7.71$ &  $58.18$ (8.4)    \\
                    & &  CART*          &  $63.33 \pm 3.36$ &  $55.17 \pm 2.16$ &  $61.47 \pm 7.13$ &  $60.00 \pm 10.33$&  $29.19 \pm 2.51$ &  $53.83$ (11.6)    \\
                    & &  FTree*     &  $38.10 \pm 0.10$ &  $61.56 \pm 2.15$ &  $51.47 \pm 17.52$&  $50.62 \pm 9.19$ &  $30.04 \pm 5.02$ &  $46.36$ (11.4)    \\
                    \cdashline{2-9}[4pt/2pt]
                    & \multirow{6}{*}{Greedy} &  VIP            &  $71.43 \pm 4.20$ &  $61.88 \pm 0.21$ &  $70.49 \pm 9.01$ &  $86.94 \pm 17.12$&  $33.53 \pm 1.32$ &  $64.86$ (3.8)    \\
                    & &  ReMLP*         &  $64.05 \pm 6.66$ &  $56.58 \pm 1.70$ &  $73.12 \pm 9.38$ &  $90.00 \pm 9.73$ &  $38.25 \pm 6.29$ &  $64.40$ (4.0)    \\
                    & &  CART*          &  $62.86 \pm 4.16$ &  $58.73 \pm 3.65$ &  $70.00 \pm 9.05$ &  $83.33 \pm 16.35$&  $37.07 \pm 7.74$ &  $62.40$ (6.4)    \\
                    & &  INVASE         &  $67.03 \pm 2.92$ &  $49.71 \pm 0.23$ &  $83.11 \pm 5.52$ &  $81.94 \pm 3.27$ &  $31.62 \pm 1.61$ &  $62.68$ (8.0)    \\
                    & &  MLP*           &  $60.24 \pm 7.88$ &  $51.85 \pm 4.24$ &  $65.25 \pm 6.07$ &  $84.45 \pm 12.91$&  $37.34 \pm 5.36$ &  $59.82$ (8.4)    \\
                    & &  FTree*         &  $51.45 \pm 5.79$ &  $56.21 \pm 0.54$ &  $68.67 \pm 8.69$ &  $84.88 \pm 8.90$ &  $29.93 \pm 2.78$ &  $58.23$ (9.0)    \\
                    & &  DIME           &  $63.22 \pm 3.34$ &  $56.55 \pm 0.25$ &  $66.39 \pm 7.96$ &  $63.06 \pm 3.22$ &  $34.65 \pm 4.83$ &  $56.77$ (8.2)    \\

        \bottomrule
    \end{tabular}
    }
    \label{tab:comparison}
\end{table*}

\begin{table}[t]            
  \centering  
  \caption{Effect of epistemic uncertainty weight $\lambda$ on feature selection performance. Results averaged across 5 datasets and 7 budget levels (20\%--80\%).}           
  \label{tab:lambda_ablation} 
  \begin{tabular}{lcc}        
  \toprule    
  $\lambda$ Group & Accuracy & Epistemic Unc. \\         
  \midrule    
  $\lambda < 0$ (penalize EU) & 66.60 $\pm$ 4.57 & 0.025 $\pm$ 0.049 \\
  $\lambda = 0$ (baseline) & 66.61 $\pm$ 3.46 & 0.009 $\pm$  0.043 \\
  $\lambda > 0$ (reduce EU) & 66.60 $\pm$ 4.41 & 0.001 $\pm$ 0.070 \\
  \bottomrule 
  \end{tabular}               
\end{table}     

\paragraph{Accuracy Comparison.} Table \ref{tab:comparison} reports the results using randomly-generated variable feature costs. Here, the proposed ReMLP with RL-based feature selection achieves the best overall performance with an average accuracy of 66.61\% across all tabular datasets. Among greedy methods, VIP achieves 64.86\%, followed closely by ReMLP-Greedy at 64.40\%. Notably, static feature selection using CMI achieves 62.33\%, remaining competitive with many dynamic methods. Rule-based approaches (CART and FTree) show different performance depending on the feature selection strategy. With greedy selection, CART achieves 62.40\%, performing comparably to static CMI and outperforming several established DFS methods, including INVASE at 62.68\% and DIME at 56.77\%. However, with RL-based selection, both rule-based methods suffer significant performance degradation.
Comparing RL and Greedy approaches, RL-based methods generally outperform their greedy counterparts when using neural architectures (ReMLP-RL: 66.61\% vs ReMLP-Greedy: 64.40\%; MLP-RL: 63.71\% vs MLP-Greedy: 59.82\%). These results also validate the effectiveness of our reparametrization strategy to reduce model risk: ReMLP consistently outperforms the standard MLP across all dataset-strategy combinations except for the Yeast dataset with greedy selection (MLP-Greedy: 37.34\% vs ReMLP-Greedy: 38.25\%). 

\paragraph{Effect of Epistemic Uncertainty in the DFS Policy.} 
We investigate incorporating epistemic uncertainty into feature selection through the weighted policy in Eq.~(\ref{eq:gain}), evaluating $\lambda$ across the $[-1.0, 1.0]$ range. Table~\ref{tab:lambda_ablation} summarises results aggregated by sign. As intended, positive $\lambda$ values reduce epistemic uncertainty compared to the baseline, while negative values increase it. However, the impact on accuracy is modest. The optimal configuration found ($\lambda = 0.1$) achieves 0.9\% relative improvement over the baseline. These findings indicate epistemic uncertainty provides a complementary signal to performance-centric objectives, with moderate gains when properly weighted.

\begin{figure}
\centering
    \includegraphics[width=.9\linewidth]{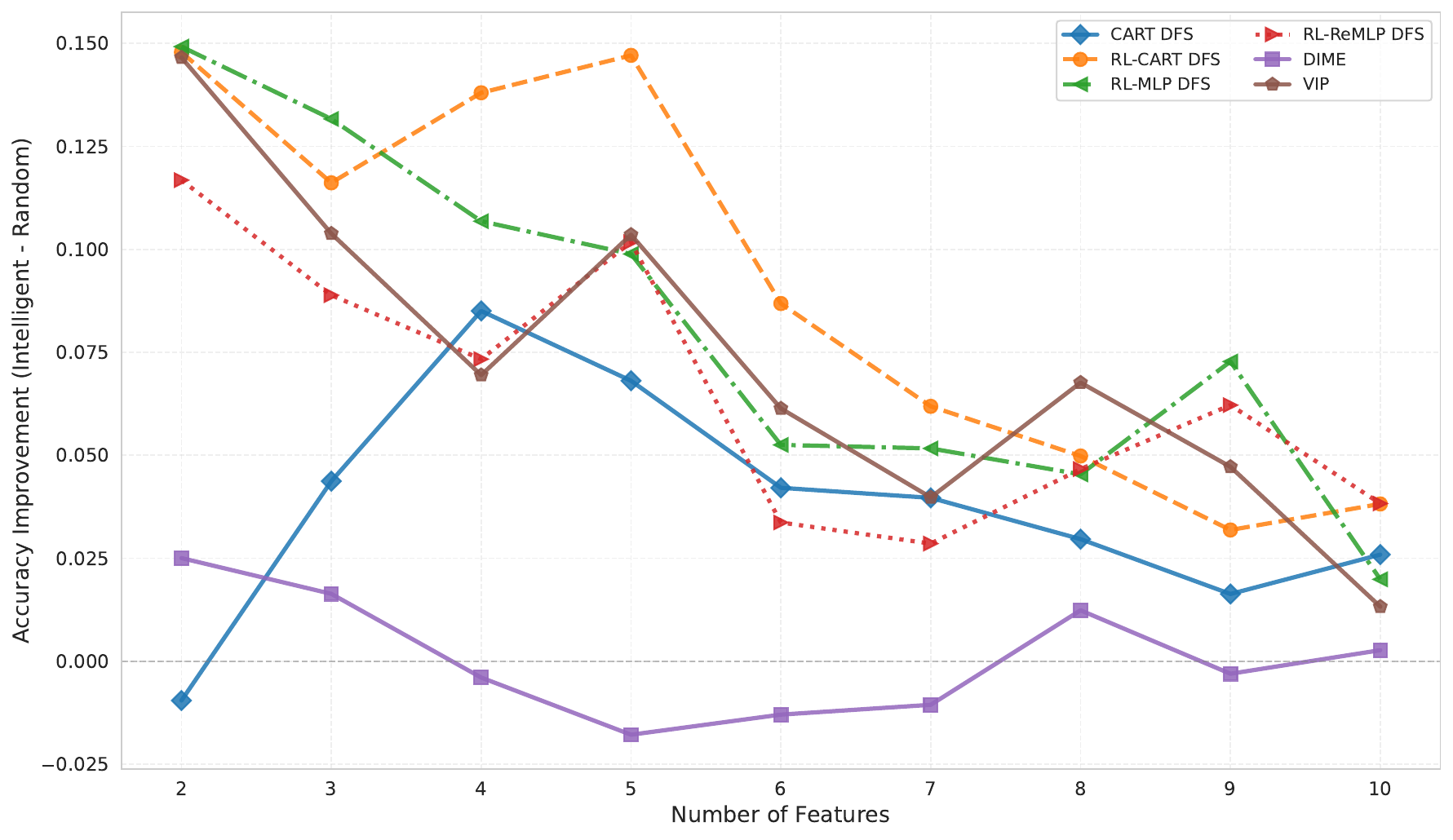}
    \caption{Evolution of average accuracy difference between feature selected and random masks for increasing budgets in the tabular datasets studied.}
    \label{fig:acc_random_analysis}
\end{figure}

\paragraph{Model adaptability and Decision Uncertainty.} Figure \ref{fig:acc_random_analysis} shows the accuracy difference between using trained feature selection policies versus random feature masks at each budget level. This analysis examines the effectiveness of the learned feature selection policy and the performance of the classifier in unseen feature subsets. As expected, we can see that some of the best performers (ReMLP, VIP) have very effective feature selection policies that consistently outperform random masks. The contrary can be seen in DIME, which shows how important it is to have a good feature selector, as DIME's average performance was the worst of all greedy performers. Our reparametrized approaches further evidence the importance of feature selection policies. Both ReMLP-RL and ReMLP-Greedy show consistent gains over random selection, with improvements ranging from 2-5\% across most budget levels. Since both methods explicitly minimise model risk through reparametrization, these gains demonstrate that intelligent feature acquisition provides benefits even when model adaptability is addressed. Regarding prediction uncertainty, we also observed that entropy generally decreases with additional features regardless of whether features are queried using random or trained policies. This confirms that predictive entropy cannot reliably discriminate between good and bad selection policies, as we formalised in Section~\ref{sec:uncertainty_in_dfs}. Additional experiments (Appendix~\ref{sec:appendix_strengthening}) 
show that epistemic uncertainty correlates positively with the subset 
risk gap $\Delta_S$ and increases under distribution shift, while 
reparametrization consistently reduces $\Delta_S$ across feature subsets.

\paragraph{Prediction Bias during Sequential Acquisition.}
A key question in DFS is whether methods correctly approximate the true conditional distribution $p(y \mid \mathbf{x}_S)$ for an observed feature subset $S$, or whether predictions implicitly rely on imputation artefacts and feature correlations. To assess this, we generate feature subsets of varying cardinalities for each dataset, train a reference classifier exclusively on $\mathbf{x}_S$ as an empirical proxy for the ground-truth conditional $p(y \mid \mathbf{x}_S)$. Then, we compare DFS predictions against this reference and a full-feature model using KL divergence. Larger divergence from the subset-specific reference, relative to the full-feature model, indicates that the method implicitly relies on imputation or feature correlation rather than learning proper conditionals. Results are shown in Table~\ref{tab:bias_summary}, averaged across six methods. DIME exhibits low divergence from the subset-adaptive reference and reduced agreement with the full model, indicating effective adaptation with limited leakage. In contrast, RL-based methods show higher agreement with the full model despite comparable or larger divergence from the subset reference, suggesting systematic reliance on imputed features. Greedy-CART-DFS incurs large divergence from both references, reflecting unstable adaptation. Notably, methods with stronger predictive performance tend to exhibit larger relative differences to the local oracles, revealing a tension between accuracy and calibration in the sequential acquisition setting.

Detailed results per dataset are available in Appendix \ref{apx:experimental_results}. More experiments testing the relationship between epistemic uncertainty, reparametrization and $\Delta_S$ are available in Appendix \ref{sec:appendix_strengthening}.

\begin{table}[ht]
\centering
\caption{
Average prediction bias metrics across datasets.
$\mathrm{KL}_{\text{Ref}}$ captures distance to subset optimal predictions, while $\mathrm{KL}_{\text{Full}}$ captures distance to whole feature set predictions.
}
\begin{tabular}{lcccc}
\toprule
Method & $\mathrm{KL}_{\text{Ref}}$
& $\mathrm{KL}_{\text{Full}}$
& KL Ratio  \\
\midrule
RL-ReMLP-DFS & 0.883 & 0.734 & 1.11  \\
RL-MLP-DFS  & 0.366 & 0.377 & 0.98 \\
VIP         & 0.393 & 0.375 & 1.04 \\
DIME        & 0.153 & 0.208 & 0.74 \\
CWCF        & 0.795 & 0.557 & 1.21 \\
Greedy-CART-DFS & 1.715 & 1.453 & 1.18 \\
\bottomrule
\end{tabular}

\label{tab:bias_summary}
\end{table}

\subsection{Results for Image Datasets Data. }
\begin{table}[]
    \centering
    \caption{Accuracy Comparison of Dynamic feature Selection methods for image datasets. Results show the average accuracy across all 1-10 patches.}
    \adjustbox{width=\linewidth}{
    \begin{tabular}{c|ccc}
    \toprule
    Method & MNIST & Cifar10 & Imagenette \\    
    \midrule
         DIME & 77.94 $\pm$ 13.64 & 55.57 $\pm$ 17.60 & 72.09 $\pm$ 10.81 \\
         VIP & 75.76 $\pm$ 19.35 & 54.20 $\pm$ 18.00 & 81.01 $\pm$ 8.90\\
         \midrule 
         DFS-CNN & 90.70 $\pm$ 8.97 & 33.37 $\pm$ 12.29 & 46.58 $\pm$ 21.76 \\
         DFS-ReCNN & 86.70 $\pm$ 15.02 &  30.47 $\pm$ 12.94 & 55.76 $\pm$ 17.17\\
         DFS-AdCNN & 95.52 $\pm$ 4.58 & 51.15 $\pm$ 12.60 & 69.25 $\pm$ 13.04\\
         DFS-AdReCNN & 96.54 $\pm$ 2.90 & 47.60 $\pm$ 14.15 & 68.08 $\pm$ 8.79\\
         \bottomrule
    \end{tabular}}
    \label{tab:res_vision_dfs_acc}
\end{table}

\paragraph{Accuracy Comparison.}
Results on image datasets differ markedly from tabular data (Table~\ref{tab:res_vision_dfs_acc}). Backbone performance dominates over the DFS policy, and VIP achieves the best overall accuracy, while approaches relying on pre-established convolutional backbones suffer substantial drops on \emph{CIFAR-10} and \emph{Imagenette}. Reparametrization alone did not yield significant improvements, suggesting that last-layer adaptation is insufficient to compensate for the mismatch between model and available information. Explicit subset adaptation (DFS-AdCNN) does improve over DFS-CNN, confirming that retraining strategies matter for partial feature visibility in image settings.

\begin{figure}
    \centering
    \includegraphics[width=\linewidth]{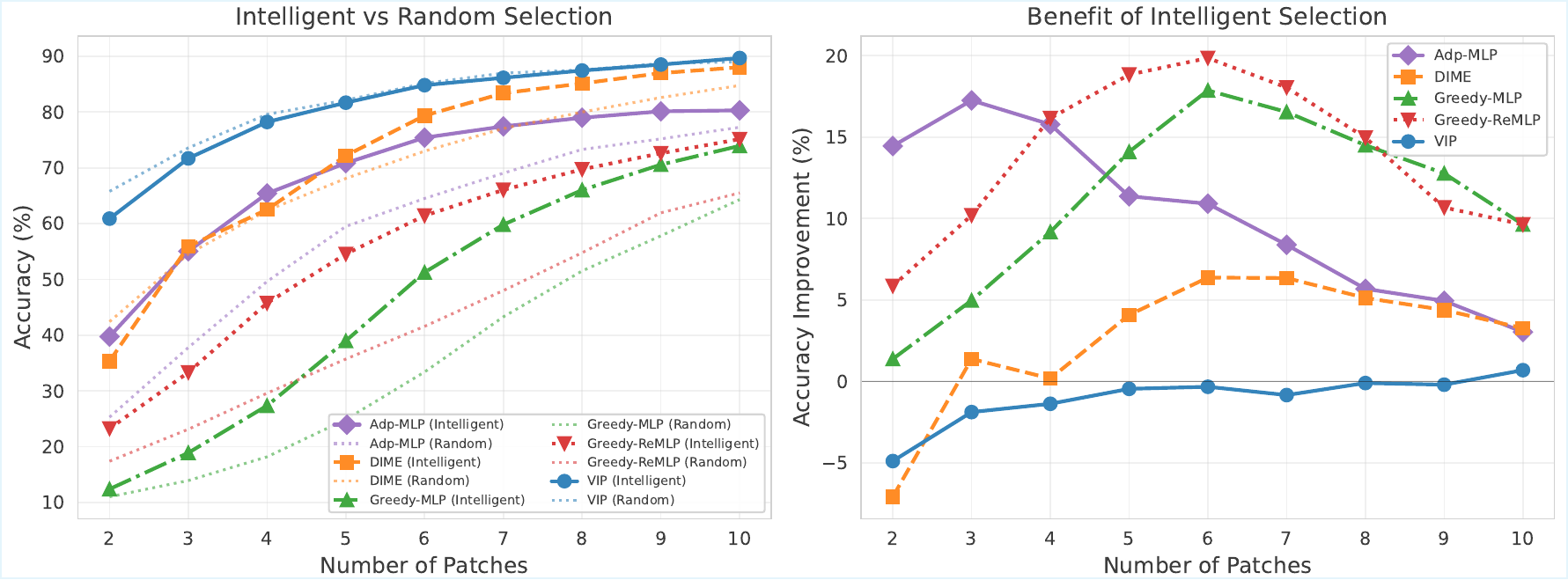}
   \caption{Comparison between intelligent and random feature acquisition on \emph{Imagenette}. Left: Classification accuracy for intelligent feature selection policies (solid lines) and random acquisition baselines (dotted lines). Right: Accuracy improvement of intelligent feature selection over random acquisition.}

    \label{fig:acc_improvement_dl}
\end{figure}

\begin{figure}
    \centering
    \includegraphics[width=\linewidth]{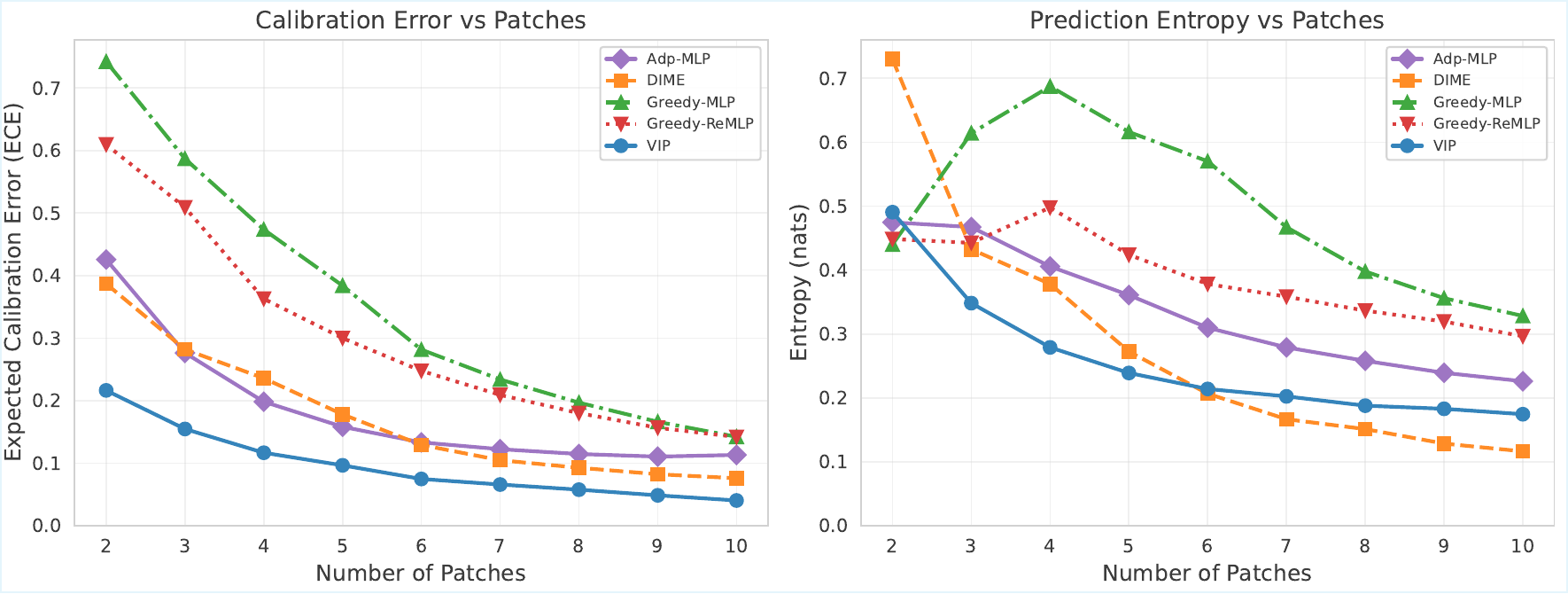}
    \caption{Calibration behaviour on \emph{Imagenette} under dynamic feature acquisition. Left: ECE as a function of the number of acquired image patches. Right: predictive entropy evolution across acquisition steps.}

    \label{fig:calibration_vision}
\end{figure}

\paragraph{Model Adaptability and Decision Uncertainty.}
Learned feature selection policies present substantial accuracy gains over random acquisition for our proposed methods, which validates the effectiveness of our DFS policy (Figure~\ref{fig:acc_improvement_dl}). In contrast, VIP exhibits virtually no gains from its DFS policy. Although surprising, this shows that a predictive backbone that is already strongly trained can overcome poorly adaptive patch selection at test time. DIME scales better at larger budgets than AdCNN, which appears to stem from a backbone better prepared to operate with increasing feature availability, as its improvement margins over random policies are smaller than in the AdCNN case.
Across all vision datasets, epistemic uncertainty remains consistently low relative to aleatoric uncertainty, making uncertainty decomposition less informative for distinguishing model risk from intrinsic data ambiguity. Calibration improves monotonically as additional features are acquired (Figure~\ref{fig:calibration_vision}) and is visibly correlated with predictive performance. This calibration evolution indicates that early cutoffs in the acquisition process cannot be safely determined using confidence thresholds alone. For example, predictions obtained with two patches using VIP and AdMLP exhibit, on average, very similar entropy levels, yet their calibration errors differ significantly.


\section{Discussion}
\paragraph{Empirical Performance Validation.}
For tabular data, RL-ReMLP achieved the best overall ranking, confirming that reparametrization improves subset-adaptive performance beyond existing DFS adaptation mechanisms, and that RL-based policies generally outperform their greedy counterparts. Among interpretable models, Greedy-CART outperformed CWCF, DIME, and INVASE despite not being designed for dynamic acquisition, validating the effectiveness of our model-agnostic DFS policy. However, the gap between rule-based and neural backbones confirms that classifier capacity remains a relevant factor in DFS performance. For image datasets, our DFS policy has significant gains over random baselines, but backbone performance dominated overall, with VIP and DIME achieving the best results. This contrast suggests two viable strategies for DFS: investing in a robust predictor that tolerates arbitrary subsets (as in VIP and DIME), or pairing an effective selector with a less accurate adaptive model. 
\paragraph{Uncertainty Quantification Challenges.}
Our empirical results support the uncertainty issues formalised in Section~\ref{sec:uncertainty_in_dfs}. Best-performing methods exhibit larger biases towards the full-information distribution $p(y \mid \mathbf{x})$ (Table~\ref{tab:bias_summary}), rather than producing well-calibrated estimates of $p(y \mid \mathbf{x}_S)$, while our reparametrization trades some predictive performance for better subset alignment. Predictive entropy decreased at similar rates under both learned and random policies, confirming that confidence cannot reliably discriminate between good and bad selection strategies. Moreover, for tabular data, not even accuracy is strictly monotonic with respect to budget (Figure~\ref{fig:acc_evo_individual_var}, Appendix \ref{apx:experimental_results}), reinforcing our concerns about confidence-based stopping criteria (Sections~\ref{sec:aleatoric_estimation} and~\ref{sec:non_monotinicy}). The apparent success of such criteria in the literature likely reflects their ability to capture decision stability rather than true class probabilities: predictions may remain unchanged as features are acquired, creating an impression of reliability even when the underlying true distribution shifts.

\paragraph{Why Static Selection Remains Competitive for Tabular Data?}
Static CMI-based feature selection achieved competitive or superior performance compared to several DFS methods. Three factors explain this. First, static selection eliminates subset-induced $\Delta_S$: the classifier is trained and evaluated on a single fixed subset, avoiding the specialisation gap from joint optimisation over exponentially many configurations. Second, it avoids the imputation bias that can appear in DFS methods, since no approximation of $p(x_{\bar S}\mid x_S)$ is required at inference time. Third, when conditional feature relevance does not vary substantially across samples, instance-specific and global CMI rankings largely coincide, limiting the benefit of adaptation while the reduced epistemic variability of static selection works in its favour.

\section{Conclusions} \label{sec:conclusions}
In this work, we formalised how DFS introduces distinct sources of uncertainty beyond the static setting: model adaptation to different feature subsets creates non-uniform epistemic uncertainty across the exponential feature space, standard imputation techniques bias intermediate estimates, and confidence-based stopping criteria cannot reliably discriminate between good and bad selection policies. These findings have direct implications for deployment in high-stakes domains, where systems must reliably quantify their uncertainty. Beyond this analysis, we introduced a model-agnostic DFS framework that adapts pre-trained classifiers, including interpretable-by-design models, to the sequential acquisition setting through subset reparametrization.

Future work should explore tailored adaptation strategies for interpretable classifiers, including extensions to non-tabular domains through concept-based architectures~\citep{koh2020concept}, as well as more expressive reparametrization techniques to further close the subset adaptation gap.

\section{Acknowledgement}

This research and Javier Fumanal-Idocin were supported by EU Horizon Europe under the Marie Skłodowska-Curie COFUND grant No 101081327 YUFE4Postdocs. Raquel Fernandez-Peralta is funded by the EU NextGenerationEU through the Recovery and Resilience Plan for Slovakia under the project No. 09I03-03-V04- 00557.

The authors acknowledge the use of the High Performance Computing Facility (Ceres) and its associated support services at the University of Essex in the completion of this work.

\bibliography{aaai2026}

\appendix
\onecolumn

\section*{Appendix Overview}

The appendix is organised into three parts.

\textbf{Theoretical Analysis.}
Appendices A–D provide additional theoretical results and methodological details:
\begin{itemize}
    \item Appendix A: Properties of the proposed epistemic uncertainty metric.
    \item Appendix B: Formal relation between prediction similarity minimisation and conditional mutual information.
    \item Appendix C: Efficient adaptation of rule-based classifiers to arbitrary feature subsets.
    \item Appendix D: Adaptation of classical rule-learning heuristics to the DFS setting.
\end{itemize}

\textbf{Extended Empirical Results.}
Appendix E reports additional experimental analyses:
\begin{itemize}
    \item Performance under uniform feature costs.
    \item Per-dataset accuracy and calibration evolution.
    \item Selective prediction and accuracy–rejection curves.
    \item Bias decomposition of sequential predictions.
\end{itemize}

\textbf{Targeted Validation Experiments.}
Appendix F reports additional experiments designed to empirically validate additional phenomena of interest:

\begin{itemize}
    \item Robustness under distribution shift.
    \item Correlation between epistemic uncertainty and subset risk gap $\Delta_S$.
    \item Effect of reparametrization on reducing subset risk.
    \item Correlation between aleatoric and epistemic uncertainties.
\end{itemize}

\textbf{Implementation Details.}
Appendix G provides reproducibility information, including dataset statistics, model architectures, hyperparameters, and training configurations.

\section{Properties of the Proposed Epistemic Uncertainty Metric}\label{apx:metric_analysis}

For uncertainty decomposition, a standard information-theoretic identity is
\begin{equation} \label{eq:uncertainty_decomposition}
H(Y\mid \mathbf{x}_S)
=
H(Y\mid \mathbf{x}_S,\Theta)
+
I(Y;\Theta\mid \mathbf{x}_S),
\end{equation}
where $\Theta$ is a random variable of model's parameters $\mathbf{\theta}\sim Q$ \citep{hullermeier2021aleatoric}. The first term is commonly interpreted as aleatoric uncertainty (AU), since it captures the irreducible uncertainty in $Y$ that remains even if the model parameters were known. The second term, $I(Y;\Theta\mid \mathbf{x}_S)$, is interpreted as epistemic uncertainty (EU), since it quantifies the expected reduction in uncertainty about $Y$ that would be obtained by observing $\Theta$.
In practice, these quantities are approximated via an ensemble of $M\in\mathbb{N}$ predictors $\{\mathbf{\theta}^{(i)}\}_{i=1}^M$ sampled from $Q$. Denoting
\begin{equation}
p_i(y)\equiv p(y\mid \mathbf{x}_S,\mathbf{\theta}^{(i)}),
\qquad
\bar p(y)\equiv \frac{1}{M}\sum_{i=1}^M p_i(y),
\end{equation}
we obtain the Monte Carlo estimators
\begin{equation}\label{eq:AU}
\mathrm{AU}(\mathbf{x}_S)
\;\approx\;
\frac{1}{M}\sum_{i=1}^M H(p_i),
\end{equation}
\begin{equation}\label{eq:EU}
\mathrm{EU}(\mathbf{x}_S)
\;\approx\;
H(\bar p)\;-\;\frac{1}{M}\sum_{i=1}^M H(p_i)
\;=\;
\frac{1}{M}\sum_{i=1}^M D_{\mathrm{KL}}(p_i\|\bar p).
\end{equation}
However, our setting differs from standard ensemble inference: predictions and explanations are produced by a single fixed model, denoted by
\begin{equation}
q(y)\equiv p(y\mid \mathbf{x}_S,\mathbf{\theta}),
\end{equation}
while auxiliary ensemble members are used only for uncertainty quantification. To measure epistemic uncertainty around the selected model, we use the quantity Eq.~(\ref{eq:epistemic_normal}), which measures how much other reasonable models would disagree with the predictor we actually use at $\mathbf{x}_S$ and thus reflects how stable that decision is under small changes in training.
It is straightforward to prove that this formula has the following relationship with standard ensemble epistemic uncertainty
\begin{equation}\label{eq:epi_gap_decomp}
e(\mathbf{x}_S) = \frac{1}{M}\sum_{i=1}^M D_{\mathrm{KL}}(p_i\|q)
=
\frac{1}{M}\sum_{i=1}^M D_{\mathrm{KL}}(p_i\|\bar p)
+
D_{\mathrm{KL}}(\bar p\|q) = EU(\mathbf{x}_s) + D_{\mathrm{KL}}(\bar p\|q).
\end{equation}
By using the definition in Eq.~(\ref{eq:epistemic_normal}) for epistemic uncertainty, we capture both the usual disagreement among ensemble members around their average prediction (left term) and the additional discrepancy introduced by relying on the single fixed model $q$ at inference time, rather than the ensemble mean $\bar p$ (right term).

Importantly, while the left term is bounded by $\ln C$ (where $C$ is the number of classes), the right term $D_{\mathrm{KL}}(\bar p\|q)$ is unbounded and can grow arbitrarily large when $q$ assigns near-zero probability to regions where $\bar p$ has significant mass. This asymmetry is particularly beneficial for uncertainty quantification: when the selected model $q$ significantly deviates from the ensemble consensus $\bar p$, the unbounded right term dominates the decomposition, effectively raising a strong alarm about potential model misspecification. In contrast, the standard epistemic uncertainty (left term alone) could remain relatively small even when $q$ is a poor representative of the ensemble, failing to capture this critical aspect on explainable models when ensemble inference is not adequate.

Now, we discuss the properties of $e(x_S)$ as a measure of epistemic uncertainty. Since it is defined via the Kullback--Leibler divergence, $e(x_S)$ is non-negative and vanishes if and only if the ensemble distribution $Q$ reduces to a Dirac measure $Q=\delta_\theta$. In this case, all ensemble members coincide, yielding $p_i=\bar p=q$ for all $i$. Thus, zero epistemic uncertainty characterises the robustness of the selected model. Nonetheless, like the mutual information approach \cite{wimmer2023quantifying}, our measure $e(x_S)$ also exhibits some non-standard behaviours: mean-preserving spreads may not increase $e(x_S)$, location shifts may not preserve it, and it is not maximal when the second-order distribution over predictions is uniform. However, these properties are less relevant in our framework, as we aim to capture epistemic uncertainty from a different perspective.  Indeed, in our case, we are not interested in minimising $EU(\mathbf{x}_S)$ alone, but rather in balancing this term with $D_{\mathrm{KL}}(\bar p\|q)$ to assess whether the selected model is coherent with the ensemble.

\begin{figure}[ht]
    \centering
    \begin{subfigure}[t]{0.48\textwidth}
        \centering
        \includegraphics[width=\linewidth]{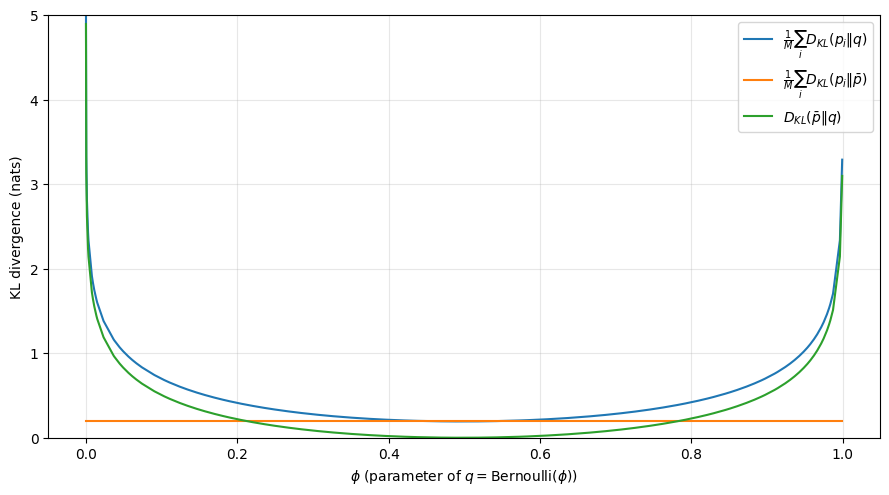}
        \subcaption{$\theta_i \sim \mathrm{Uniform}(0,1)$}
        \label{fig:epistemic-uniform}
    \end{subfigure}
    \hfill
    \begin{subfigure}[t]{0.48\textwidth}
        \centering
        \includegraphics[width=\linewidth]{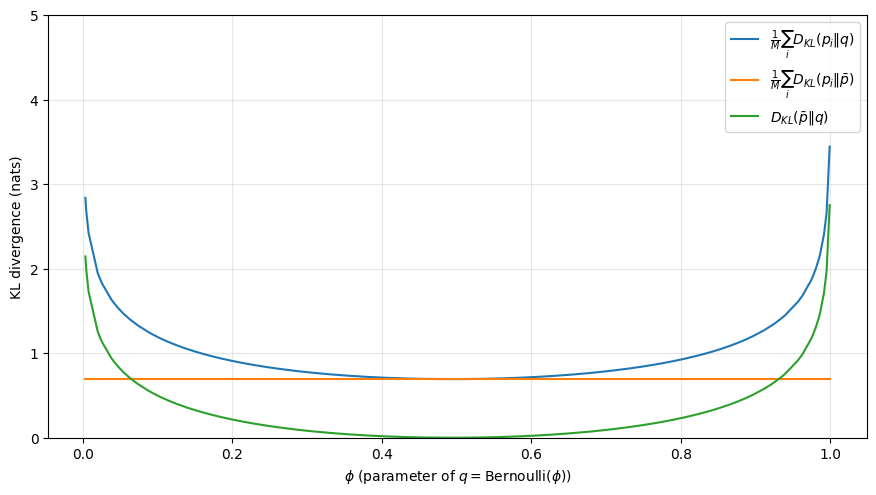}
        \subcaption{$\theta_i \sim \tfrac12\delta_0 + \tfrac12\delta_1$}
        \label{fig:epistemic-atomic}
    \end{subfigure}
    \caption{Epistemic uncertainty decomposition for Bernoulli ensembles.}
    \label{fig:epistemic-gap}
\end{figure}

To illustrate the behaviour of our measure, we consider the following experiment. We study an ensemble of Bernoulli models $p_i=\mathrm{Bernoulli}(\theta_i)$ and compute $e(x_S)$ for different reference models $q=\mathrm{Bernoulli}(\phi)$. We compare two choices for the distribution $Q$ of the parameters $\theta_i$: (i) a uniform distribution on $[0,1]$, and (ii) an atomic distribution $Q=\tfrac12\delta_0+\tfrac12\delta_1$. Fig.~\ref{fig:epistemic-gap} compares the behaviour of the $e(\mathbf{x}_S)$ decomposition for these two choices of the second-order distribution $Q$. As shown in \cite{wimmer2023quantifying}, the term $E(\mathbf{x}_S)$ is close to its theoretical maximum $\ln(2)\approx 0.69$ for the mixture of Dirac distributions, while for the uniform distribution this term is much smaller, around $0.19$. For the second term, $D_{\mathrm{KL}}(\bar p\|q)$, we observe that it becomes large at the extremes, where the selected model is very different from the ensemble average. When the selected model is more consistent with the ensemble, this term is close to zero, and only the ensemble epistemic uncertainty remains. This behaviour is aligned with our objective, as the measure naturally penalises incoherent model choices while remaining stable when the selected model agrees with the ensemble.

\section{Relation between Prediction Similarity minimization and CMI}\label{appendix:CMIcomparison}

In this section, we study the relationship between the conditional mutual information maximisation and the prediction difference minimisation as DFS policies.

First, we rewrite the decrease in prediction discrepancy obtained by adding a candidate feature $x_i$ to the currently observed set $S$.

\begin{lemma}\label{lemma:apA:1}
Let $u(\mathbf{x}_S) = D_{\mathrm{KL}}(p(\hat{y}\mid \mathbf{x}) \,\|\, p(\hat{y}\mid \mathbf{x}_S))$ and define
\[
\Delta u_i = u(\mathbf{x}_S) - u(\mathbf{x}_S \cup \{x_i\}).
\]
Then
\[
\Delta u_i
=
\sum_{\hat{y}} p(\hat{y}\mid \mathbf{x})
\log\left(
\frac{p(\hat{y}\mid \mathbf{x}_S,x_i)}{p(\hat{y}\mid \mathbf{x}_S)}
\right).
\]
\end{lemma}

\begin{proof}
By definition,
\begin{align*}
u(\mathbf{x}_S)
&=
\sum_{\hat{y}} p(\hat{y}\mid \mathbf{x})
\log\frac{p(\hat{y}\mid \mathbf{x})}{p(\hat{y}\mid \mathbf{x}_S)},\\
u(\mathbf{x}_S \cup \{x_i\})
&=
\sum_{\hat{y}} p(\hat{y}\mid \mathbf{x})
\log\frac{p(\hat{y}\mid \mathbf{x})}{p(\hat{y}\mid \mathbf{x}_S,x_i)}.
\end{align*}
Subtracting yields
\[
\Delta u_i
=
u(\mathbf{x}_S) - u(\mathbf{x}_S \cup \{x_i\})
=
\sum_{\hat{y}} p(\hat{y}\mid \mathbf{x})
\log\frac{p(\hat{y}\mid \mathbf{x}_S,x_i)}{p(\hat{y}\mid \mathbf{x}_S)}.
\]
\end{proof}

Next, we compare the conditional mutual information score $I(x_i;\hat{y} |\mathbf{x}_S)$ with the expected decrease in discrepancy $\mathbb{E}_{p(x_i\mid \mathbf{x}_S)}[\Delta u_i]$. The lemma below gives an exact identity, disclosing that their differences depend on how close the partial predictor $p(\hat{y}\mid \mathbf{x}_S,x_i)$ is to the global model $p(\hat{y}\mid \mathbf{x})$.

\begin{lemma}\label{lemma:apA:2}
    \begin{align*}
I(x_i;\hat{y}\mid \mathbf{x}_S) - \mathbb{E}_{p(x_i\mid \mathbf{x}_S)}[\Delta u_i]
&=
\mathbb{E}_{p(x_i\mid \mathbf{x}_S)}\left[\sum_{\hat{y}}\big(p(\hat{y}\mid \mathbf{x}_S,x_i)-p(\hat{y}\mid \mathbf{x})\big)\log\frac{p(\hat{y}\mid \mathbf{x}_S,x_i)}{p(\hat{y}\mid \mathbf{x}_S)}\right].
\end{align*}
\end{lemma}

\begin{proof}
    From Lemma~\ref{lemma:apA:1},
\[
\Delta u_i
=
\sum_{\hat{y}} p(\hat{y}\mid \mathbf{x})
\log\frac{p(\hat{y}\mid \mathbf{x}_S,x_i)}{p(\hat{y}\mid \mathbf{x}_S)}.
\]
The conditional mutual information can be written as
\[
I(x_i;\hat{y}\mid \mathbf{x}_S)
=
\sum_{x_i,\hat{y}} p(x_i,\hat{y}\mid \mathbf{x}_S) \log\frac{p(\hat{y}\mid \mathbf{x}_S,x_i)}{p(\hat{y}\mid \mathbf{x}_S)}
=
\mathbb{E}_{p(x_i\mid \mathbf{x}_S)}\left[\sum_{\hat{y}} p(\hat{y}\mid \mathbf{x}_S,x_i) \log\frac{p(\hat{y}\mid \mathbf{x}_S,x_i)}{p(\hat{y}\mid \mathbf{x}_S)}\right].
\]
Therefore,
\[
\mathbb{E}_{p(x_i\mid \mathbf{x}_S)}[\Delta u_i]
=
\mathbb{E}_{p(x_i\mid \mathbf{x}_S)}\left[\sum_{\hat{y}} p(\hat{y}\mid \mathbf{x}) \log\frac{p(\hat{y}\mid \mathbf{x}_S,x_i)}{p(\hat{y}\mid \mathbf{x}_S)}\right],
\]
and taking the difference yields the result.
\end{proof}
Notice that both criteria involve the same log-ratio term
\(
\log\frac{p(\hat{y}\mid \mathbf{x}_S,x_i)}{p(\hat{y}\mid \mathbf{x}_S)},
\)
but they differ in the weighting distribution over $\hat{y}$: mutual information
weights by $p(\hat{y}\mid \mathbf{x}_S,x_i)$, whereas the prediction-difference
gain weights by the global predictor $p(\hat{y}\mid \mathbf{x})$.
Therefore, prediction-difference minimisation can be interpreted as a targeted
version of conditional mutual information, where feature selection is guided
by alignment with a reference (global) predictive model.

The two DFS policies coincide whenever
\(
p(\hat{y}\mid \mathbf{x}) \approx p(\hat{y}\mid \mathbf{x}_S,x_i),
\)
in which case the term in Lemma~\ref{lemma:apA:2} vanishes. This situation
arises, for instance, when $S \cup \{i\}$ captures nearly all predictive
information in $\mathbf{x}$, so that the partial and global predictors agree.
More generally, the two criteria need not coincide; however, when observing
$x_i$ given $\mathbf{x}_S$ yields a low-entropy (nearly deterministic)
posterior over $\hat{y}$, both scores typically rank such a feature highly,
since the log-ratio term becomes large in magnitude and is dominated by a
single outcome.

\section{Rule-Based Model Adaptation for Arbitrary Feature Subsets}
\label{appendix:A}

Rule-based classifiers offer a uniquely efficient mechanism for adaptation to arbitrary feature subsets. Unlike neural networks that require retraining or forward passes through modified architectures, rule-based models can be reparametrized through structural manipulation and confidence re-estimation. Moreover, the explicit feature dependencies encoded in rule antecedents allow us to constrain the feature search space to only those features that can influence the final prediction.

\subsection{Rule-Based Classifier Structure}

A rule-based classifier consists of a set of rules $\mathcal{R} = \{r_1, \ldots, r_L\}$, where each rule $r$ has the form:
\begin{equation}
r: r_A \Rightarrow c,
\end{equation}
with $r_A$ being a conjunction of conditions on features (the antecedent) and $c \in \{1,\ldots,C\}$ being the predicted class (the consequent). For a sample $\mathbf{x}$, a rule fires if all conditions in $r_A$ are satisfied. The classifier's prediction is typically determined by the rule with the highest confidence or firing strength among all activated rules.

\subsection{Adaptation Procedure for Subset $S$}

Given a global rule-based model trained on the complete feature set and a subset $S$ of observed features, we adapt the model through the following steps:

\begin{enumerate}
    \item Condition Removal: for each rule $r \in \mathcal{R}$, we remove all conditions in $r_A$ that involve features in $\bar{S}$ (the unobserved features). This yields a reduced antecedent $r_A^S$ containing only conditions on observed features.
    \item Rule Pruning: Remove any rules where $r_A^S = \emptyset$ (i.e., no conditions remain after feature removal). 
    \item Confidence Re-estimation: For each remaining rule $r$ with reduced antecedent $r_A^S$, re-estimate its class prediction confidence. For crisp rules, the confidence is:
    \begin{equation}
    \text{Confidence}(r^S \Rightarrow c) =
    \frac{
        \sum_{i:\,\mathbf{x}^{(i)} \in X_{r_A^S}}
        \mathbb{I}\!\left(y^{(i)} = c\right)
    }{
        \left|X_{r_A^S}\right|
    },
    \end{equation}
    where $X_{r_A^S}$ is the set of training samples satisfying the reduced antecedent $r_A^S$.
    For fuzzy rules with membership function $\mu_r$, the confidence becomes:
    \begin{equation}
    \text{Confidence}(r^S \Rightarrow c) =
    \frac{
        \sum_{i:\,\mathbf{x}^{(i)} \in X}
        \mu_{r^S}(\mathbf{x}^{(i)}_S)
        \cdot
        \mathbb{I}(y^{(i)} = c)
    }{
        \sum_{i:\,\mathbf{x}^{(i)} \in X}
        \mu_{r^S}(\mathbf{x}^{(i)}_S)
    }.
    \end{equation}
    This re-estimation ensures predictions are unbiased estimates of $p(y|\mathbf{x}_S)$ rather than incorrectly treating missing conditions as satisfied.
\end{enumerate}

\textbf{Computational Efficiency Through Caching.}
The key advantage of this approach is that all confidence values can be precomputed during training. 
For a rule with $k$ conditions, there are $2^k$ possible sub-rules, corresponding to all subsets of its antecedent conditions. 
If we restrict rules to contain at most $k_{\max}$ conditions, the number of sub-rules per rule is therefore bounded by $2^{k_{\max}}$.

For $k_{\max} = 4$ (a reasonable constraint for interpretability), each rule generates at most 16 sub-rules. 
Even for a rule base with 100 rules, this yields only 1,600 confidence values to cache.

At inference time, adapting to any subset $S$ requires only:
\begin{enumerate}
\item Identifying which conditions remain in each rule (constant time with proper indexing).
\item Looking up the precomputed confidence for each reduced rule (constant time hash table lookup).
\end{enumerate}

\subsection{Discarding Features in the Search Space}

The output of a rule-based classifier is determined by
\[
\hat{y} = c, \quad \text{where } r^* = \arg \max_{r \in \mathcal{R}} r(\mathbf{x}),
\]
where $r(\mathbf{x})$ denotes the evaluation of rule $r$, and $r_A(\mathbf{x})$ denotes the evaluation of its antecedent. 
For crisp rules, $r_A(\mathbf{x}) \in \{0,1\}$ indicates whether all conditions are satisfied, while for fuzzy rules $r_A(\mathbf{x}) \in [0,1]$ is obtained by aggregating condition memberships using a t-norm.

Due to the conjunctive structure of rule antecedents, if any condition evaluates to zero, the antecedent evaluation becomes zero, i.e.,
\[
r_A(\mathbf{x}) = 0 \quad \text{if } \exists\, t \in r_A \text{ such that } t(\mathbf{x}) = 0.
\]
Under partial feature observation, after removing conditions involving unobserved features, the same property applies to the reduced antecedent $r_A^S$ and its evaluation $r_A^S(\mathbf{x}_S)$.

This property enables efficient search space reduction through two mechanisms:
\begin{enumerate}
    \item Early termination: if any evaluated condition returns $0$, the antecedent evaluation becomes $0$, allowing us to discard rule $r$ without evaluating the remaining conditions. More generally, a pruning threshold $\theta > 0$ can be used, which is particularly useful in fuzzy rule systems.
    
    \item Feature selection: the active feature space can be reduced to
    \[
    S_{\text{active}} =
    \bigcup_{\substack{r \in \mathcal{R} \\ r_A^S(\mathbf{x}_S) > \theta}}
    \bigcup_{t \in r_A^S}
    \mathrm{feature}(t),
    \]
    where $\mathrm{feature}(t)$ denotes the index of the feature involved in condition $t$.
\end{enumerate}

This optimisation reduces computational complexity while preserving exact rule evaluation when $\theta = 0$, and introduces a controllable approximation otherwise.

    

\subsection{Example: Decision Tree Adaptation}

Consider a decision tree (CART) as a special case of a rule-based classifier, where each leaf node corresponds to a rule. A path from root to leaf defines the rule antecedent as a conjunction of all split conditions along that path.

For a tree with maximum depth $d$, each leaf has at most $d$ conditions. The number of possible reduced rules per leaf is therefore at most $2^d$. For $d=5$, this is 32 sub-rules per leaf. A tree with 50 leaves thus requires caching at most 1,600 confidence values. Additionally, sub-rules can be shared across multiple original rules, allowing for additional space optimisation.

\section{Adapting classical heuristics in rule-learning for DFS} \label{apx:dfs_rules_adaption_naive}

Training rule-based classifiers in the DFS setting requires modifying standard split selection criteria to account for feature availability uncertainty. We describe how to adapt the greedy rule-learning algorithms used in this paper.

The aim in CART for classification is to find the split $\theta$ (on feature $m$ and threshold $t$) that maximises the purity gain, typically measured by the decrease in Gini Impurity. At a node, the Gini Impurity is defined as:
\begin{equation}
    I_G(X) = 1 - \sum_{k=1}^{K} p_k^2
\end{equation}
where $p_k$ is the proportion of class $k$ in $X$. The quality of a split $\theta$ that partitions $X$ into $X_{\text{left}}$ and $X_{\text{right}}$ is measured by the weighted impurity reduction:
\begin{equation} \label{eq:cart_split}
    \Delta I_G(X, \theta) = I_G(X) - \left( \frac{|X_{\text{left}}|}{|X|} I_G(X_{\text{left}}) + \frac{|X_{\text{right}}|}{|X|} I_G(X_{\text{right}}) \right).
\end{equation}

The optimal split is then selected as:
\begin{equation}
    \theta^* = \operatorname*{arg\,max}_{\theta} \Delta I_G(X, \theta).
\end{equation}

In the DFS setting, we must account for the fact that at inference time, only a subset $S$ of features may be observed. To train a tree that performs well across arbitrary feature subsets, we modify the split criterion to maximise the expected impurity reduction over the distribution of feature subsets that will include the splitting feature. Specifically, when considering a split on feature $j$, we compute:
\begin{equation}
    \theta^*_{\text{DFS}} =
    \operatorname*{arg\,max}_{\theta:\,\mathrm{feature}(\theta)=j}
    \mathbb{E}_{S \sim p(S \mid j \in S)}
    \left[
        \Delta I_G(X_S, \theta)
    \right],
\end{equation}
where $p(S \mid j \in S)$ is the distribution over feature subsets conditioned on $j$ being observed, and $X_S = \{(\mathbf{x}^{(i)}_S, y^{(i)})\}_{i=1}^N$ denotes the dataset where only features in $S$ are visible (other features imputed or masked). In practice, this expectation is approximated by sampling $K$ feature subsets $\{S_1, \ldots, S_K\}$ such that $j \in S_i$ for all $i$, computing $\Delta I_G(X_{S_i}, \theta)$ for each subset, and taking the empirical average $\frac{1}{K} \sum_{i=1}^K \Delta I_G(X_{S_i}, \theta)$. This approach ensures that the selected splits remain effective even when only partial information is available at test time.

For Fuzzy Tree, the adaptation is analogous.

\section{Extended Empirical Results}
\label{apx:experimental_results}

In this section, we display additional results regarding tabular data, disaggregated by individual datasets for:

\begin{itemize}
    \item Uniform feature costs.
    \item Accuracy and calibration error.
    \item Uncertainty evolution with increasing budgets.
    \item Prediction bias during feature acquisition steps.
\end{itemize}


To measure calibration, we analyse the Expected Calibration Error (ECE) for different budgets. ECE quantifies the alignment between predicted confidence and actual accuracy by partitioning predictions into equally-spaced confidence bins and measuring discrepancies:
\begin{equation}
\text{ECE} = \sum_{m=1}^{M} \frac{|B_m|}{n} |\text{acc}(B_m) - \text{conf}(B_m)|,
\end{equation}
where $B_m$ represents the $m$-th confidence bin, $n$ is the total number of samples, $\text{acc}(B_m)$ is the empirical accuracy within the bin, and $\text{conf}(B_m)$ is the average predicted confidence \cite{naeini2015obtaining}. 

\subsection{Results for uniform feature costs}
\begin{table*}[ht]
    \centering
    \caption{Accuracy comparison of Static and Dynamic Feature Selection methods for tabular datasets. Results show the average accuracy across all budget levels (1-10 features), with standard deviation indicating variability across budget levels. The final rank is computed as the mean of the ranks obtained across the five individual datasets. Methods marked with * are our proposed approaches.}
    \adjustbox{width=\linewidth}{
    \begin{tabular}{ccc|ccccc|c}
    \toprule
         &  & Method          &  Cirrhosis           &  Diabetes            &  Heart               &  Wine                &  Yeast               &  Average (rank)   \\
    \midrule
    \multirow{2}{*}{}  
&  & Static CMI            &  $69.71 \pm 2.07$ &  $62.64 \pm 0.26$ &  $79.78 \pm 3.07$ &  $95.37 \pm 4.17$ &  $55.44 \pm 5.49$ &  $72.59$ (4.2)    \\
 &              &  TabNet         &  $66.80 \pm 9.61$ &  $53.61 \pm 4.40$ &  $83.97 \pm 3.26$ &  $86.73 \pm 9.08$ &  $56.08 \pm 4.56$ &  $69.44$ (7.0)    \\
    \midrule
    \multirow{12}{*}{Dynamic} & \multirow{5}{*}{RL} &  ReMLP*         &  $70.50 \pm 3.08$ &  $58.31 \pm 1.01$ &  $80.88 \pm 2.96$ &  $98.15 \pm 3.67$ &  $54.51 \pm 1.85$ &  $72.47$ (3.8)    \\
                & &  MLP*           &  $72.49 \pm 1.10$ &  $55.01 \pm 0.41$ &  $82.88 \pm 4.19$ &  $97.84 \pm 3.04$ &  $54.51 \pm 3.18$ &  $72.55$ (4.2)    \\
                & &  CART*          &  $67.59 \pm 3.56$ &  $52.60 \pm 1.55$ &  $78.14 \pm 3.67$ &  $91.36 \pm 2.58$ &  $57.32 \pm 2.40$ &  $69.40$ (7.0)    \\
                & &  Ftree*     &  $76.32 \pm 0.93$ &  $61.36 \pm 0.14$ &  $67.21 \pm 3.29$          &  $90.74 \pm 3.10$ &  $47.51 \pm 0.59$ &  $68.63$ (7.2)    \\
                & &  CWCF           &  $74.60 \pm 5.19$ &  $53.66 \pm 0.75$ &  $76.14 \pm 13.86$&  $83.33 \pm 13.82$&  $51.37 \pm 11.05$&  $67.82$ (8.0)    \\
                \cdashline{2-9}[4pt/2pt]
                & \multirow{7}{*}{Greedy} &  VIP            &  $75.40 \pm 1.33$ &  $61.74 \pm 0.24$ &  $84.70 \pm 4.41$ &  $96.29 \pm 4.17$ &  $56.68 \pm 2.50$ &  $74.96$ (2.0)    \\
                & &  ReMLP*         &  $69.58 \pm 2.60$ &  $54.83 \pm 1.41$ &  $70.86 \pm 9.30$ &  $84.57 \pm 7.75$ &  $52.41 \pm 6.40$ &  $66.45$ (8.6)    \\
                & &  DIME           &  $65.87 \pm 3.57$ &  $55.87 \pm 0.46$ &  $72.50 \pm 9.01$ &  $79.01 \pm 10.12$&  $46.35 \pm 2.98$ &  $63.92$ (10.2)    \\
                & &  INVASE         &  $69.18 \pm 1.83$ &  $48.70 \pm 0.28$ &  $79.78 \pm 2.59$ &  $78.71 \pm 5.55$ &  $33.00 \pm 2.23$ &  $61.87$ (10.6)    \\
                & &  MLP*           &  $67.06 \pm 5.12$ &  $52.56 \pm 3.39$ &  $79.05 \pm 5.49$ &  $76.54 \pm 16.26$&  $49.98 \pm 8.06$ &  $65.04$ (10.6)    \\
                & &  CART*          &  $59.13 \pm 5.16$ &  $49.79 \pm 0.61$ &  $68.85 \pm 8.56$ &  $92.28 \pm 13.01$&  $50.47 \pm 7.70$ &  $64.10$ (10.6)    \\
                & &  FTree*         &  $53.57 \pm 7.74$ &  $54.70 \pm 0.73$ &  $71.22 \pm 0.86$ &  $87.34 \pm 11.12$&  $32.77 \pm 4.12$ &  $59.92$ (11.0)    \\
    \bottomrule
    \end{tabular}
    }
    \label{tab:comparison_uniform}
\end{table*}

Table~\ref{tab:comparison_uniform} presents the average performance and standard deviation across all budget levels (1-10 features) for all methods and datasets for uniform feature costs. RL-based methods show strong overall performance. Among RL-based approaches, both MLP and ReMLP achieve the best results (72.55\% and 72.47\%). Both rule-based approaches outperformed CWCF (67.82\%, rank 8.0) in this case, which is contrary to the results obtained with variable feature costs. Greedy optimisation approaches show surprisingly variable performance. VIP achieves the best overall results (74.96\%, rank 2.0), and was the only method outperforming static CMI selection in terms of average accuracy (72.59\%, rank 4.2). However, the rest of the greedy strategies usually perform worse than RL approaches, which is particularly visible in methods tested using both approaches.

Despite the theoretical advantages of DFS, static CMI-based feature selection (72.59\%, rank 4.2) outperforms six of the twelve dynamic methods tested, including several sophisticated approaches like DIME and INVASE. This result underscores the difficulty of learning effective DFS systems, which can outweigh the potential benefits of sample-specific feature selection. TabNet (69.44\%, rank 7.0) did fall as well below Static CMI, showing the importance of model adaptability even when all features are visible.

All methods are showing consistently better results than in the uniform case than in the variable case, as the number of total features queried is larger than in the static case. However, the different rank results also suggest that the heuristic of dividing the usefulness of a feature by its cost is more successful in some methods than others.

\subsection{Performance evolution for Individual Datasets}

\begin{figure}[ht]
    \centering
    \includegraphics[width=\linewidth]{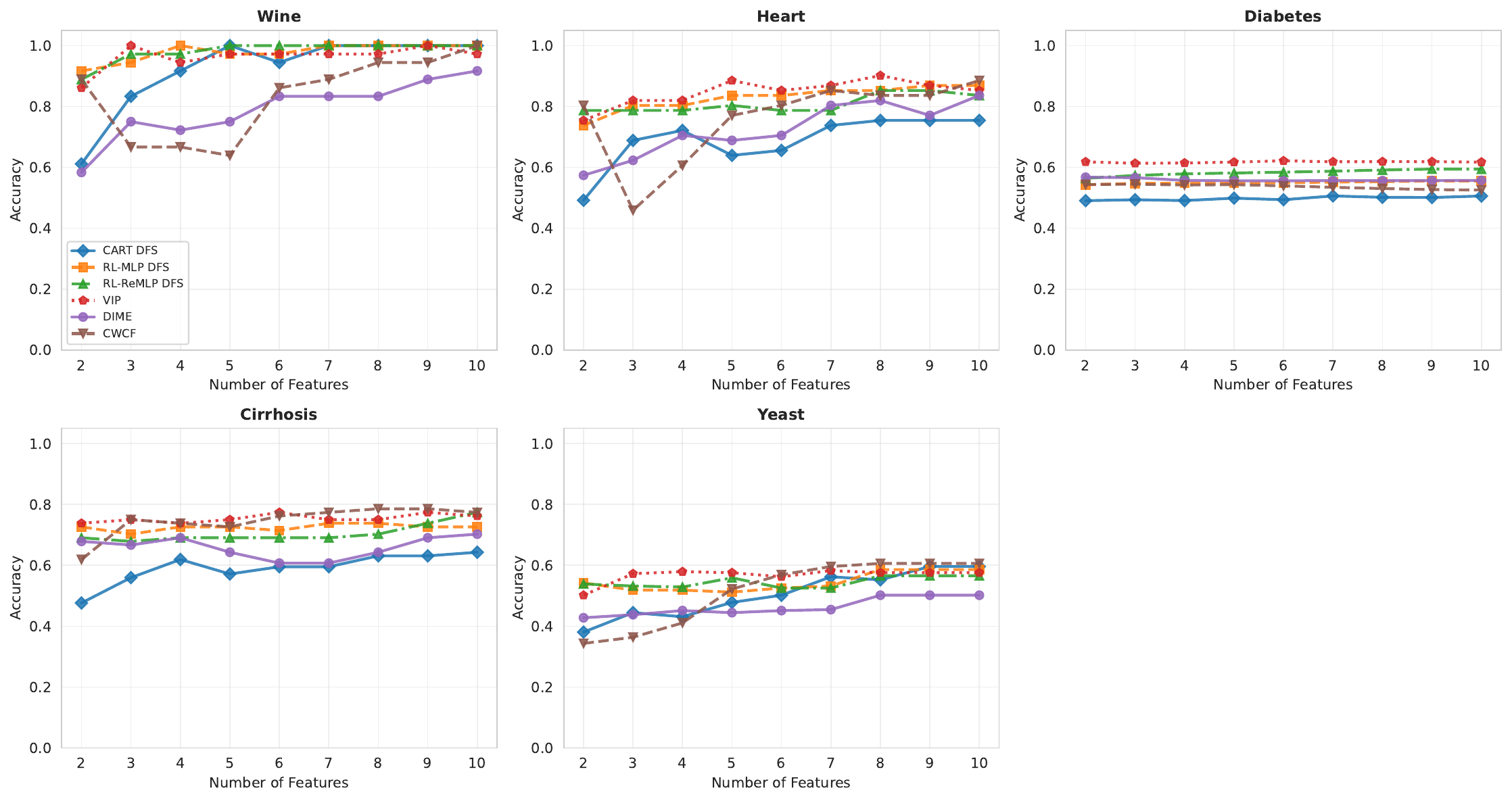}
    \caption{Accuracy evolution according to budget for all tabular datasets tested for different DFS selection methods using variable feature costs.}
    \label{fig:acc_evo_individual_var}
\end{figure}

\begin{figure}[ht]
    \centering
    \includegraphics[width=\linewidth]{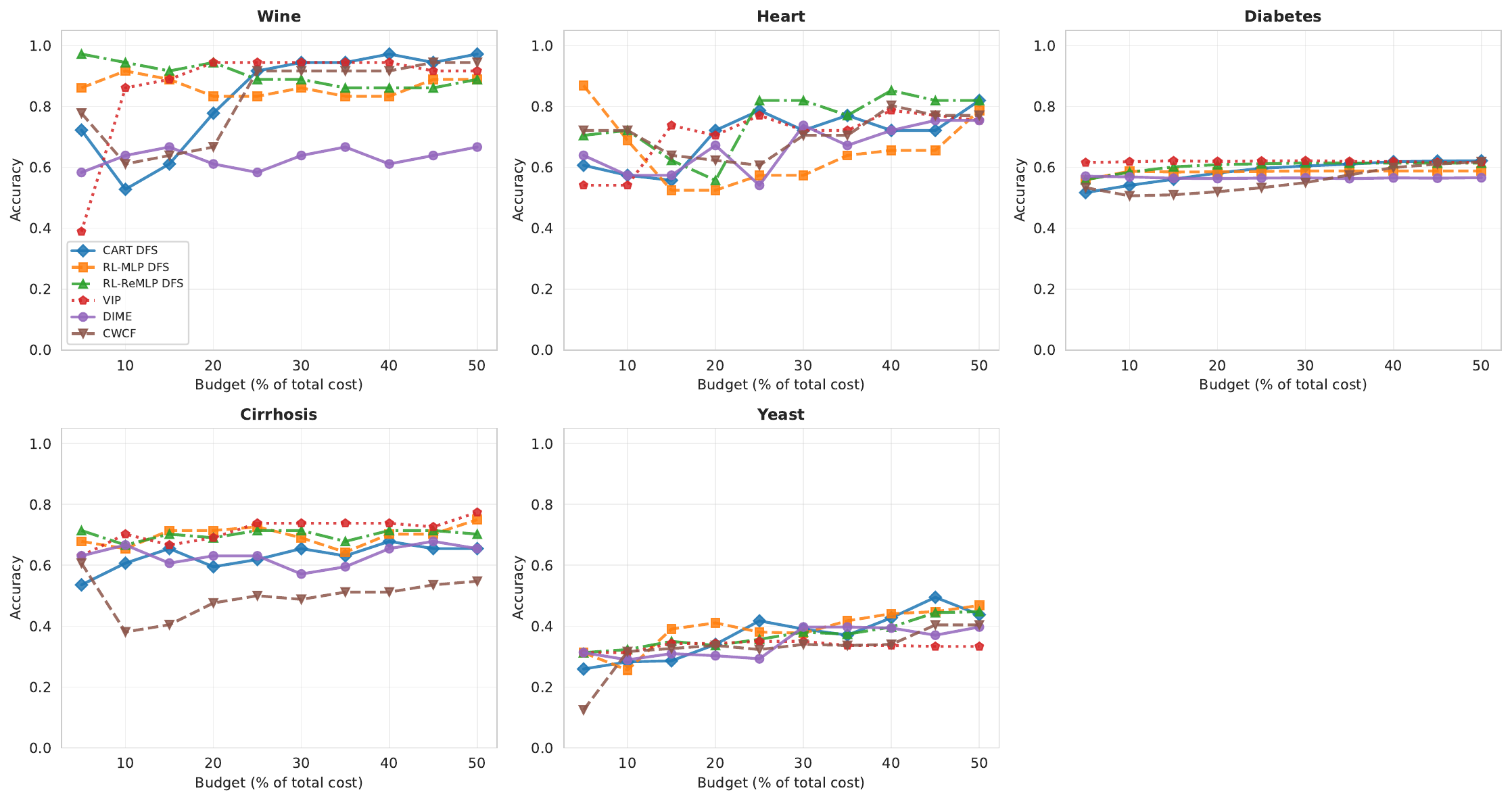}
    \caption{Accuracy evolution according to budget for all tabular datasets tested for different DFS selection methods using uniform feature costs.}
    \label{fig:acc_evo_individual_uniform}
\end{figure}

\begin{figure}[ht]
    \centering
    \includegraphics[width=\linewidth]{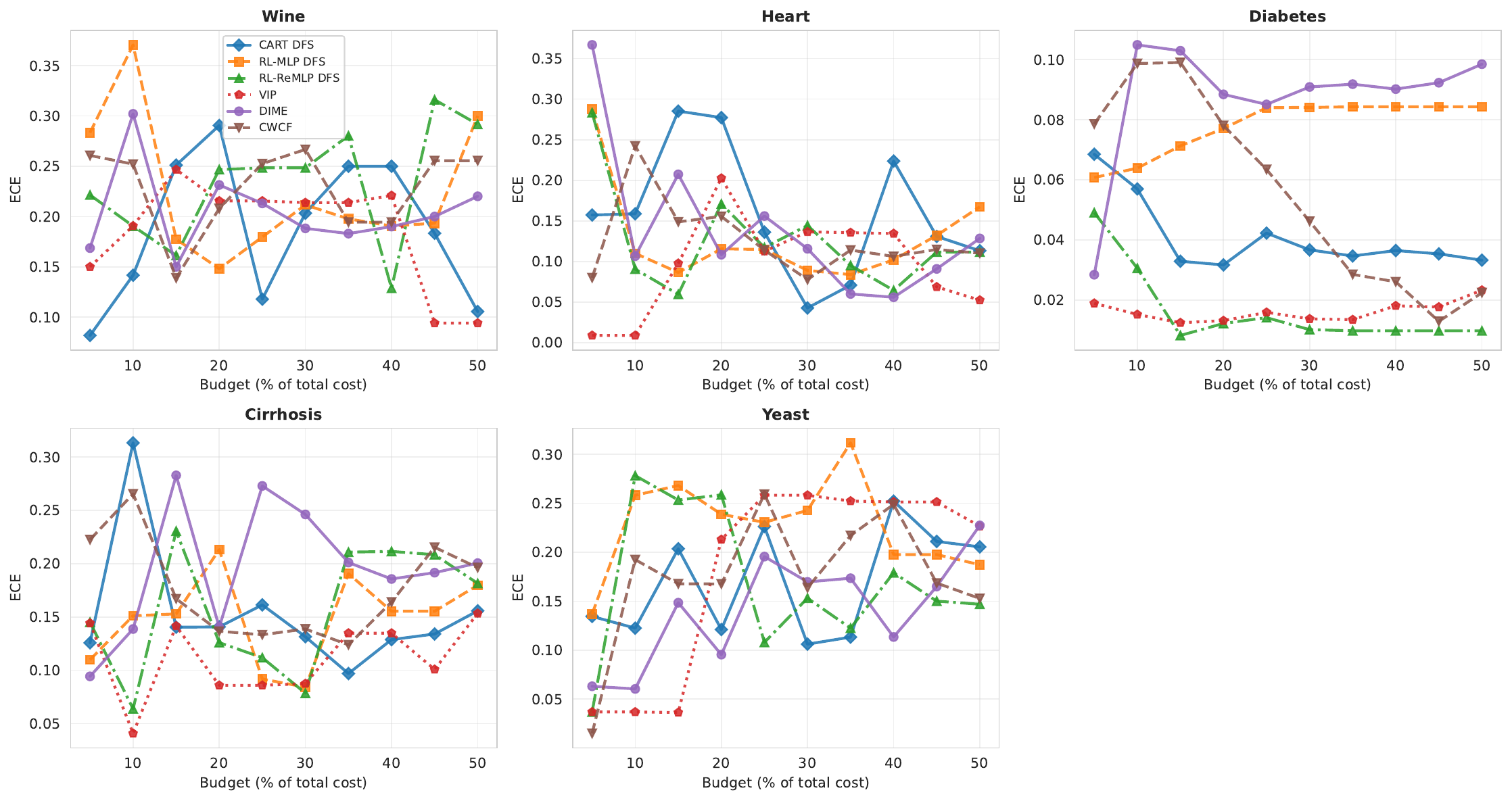}
    \caption{ECE evolution according to budget for all tabular datasets tested for different DFS selection methods using variable feature costs.}
    \label{fig:ece_evo_individual_uniform}
\end{figure}

Here, we analyse accuracy progression (Figures \ref{fig:acc_evo_individual_var} and \ref{fig:acc_evo_individual_uniform}) and calibration quality via ECE (Figure \ref{fig:ece_evo_individual_uniform}) individually per dataset. Because of visibility limitations, the figures only show the following DFS methods:  Greedy CART, RL MLP, RL-ReMLP, VIP, DIME, CWCF.

\textbf{Wine dataset.} It represents a well-structured classification problem where features have clear discriminative power. All methods show strong convergence to high accuracy: most methods achieve >95 accuracy by 7-8 features (Figure 4), with VIP, RL-ReMLP, and RL-MLP reaching near-perfect performance (98-100\%). This suggests the dataset has highly informative features that enable reliable classification with relatively small feature subsets.  However, CART-DFS shows notable volatility in the 2-4 feature range (60-85\% accuracy), only stabilising after acquiring 5+ features. This reflects the challenge of adapting decision tree structures trained for the whole set of features to very small feature subsets, which did not happen in RL-MLP, which is another method that uses a global classifier as a backbone. DIME exhibits the poorest performance in the uniform setting, but was significantly better in variable feature costs. The discrepancy between the performance in these two settings is not completely clear, but we believe that subset risk minimisation might have got stuck in some local minimum.
The ECE analysis reveals that high accuracy does not translate to reliable confidence estimates. Most methods maintain ECE of 0.15-0.26, indicating systematic overconfidence. VIP shows particularly erratic calibration (ECE ranging 0.05-0.38), with the worst calibration occurring at 4 features (ECE = 0.38) despite reasonable accuracy at that point. RL-ReMLP maintains the most stable and best calibration (ECE ~0.05 throughout). However, these results vary significantly across datasets.

\textbf{Heart dataset.} It presents a binary classification problem with moderate sample size and mixed feature informativeness. Unlike Wine, the Heart dataset shows a smoother performance evolution throughout the entire acquisition process. However, multiple methods exhibit non-monotonic accuracy curves in the uniform setting, which highlights the challenges discussed in Section 2. We can also see the importance of the backbone classifier, as most neural methods converged to the same accuracy value at maximum budget, while the CART classifier was significantly lower. ECE evolution also shows extreme volatility on Heart. No method maintains consistently good calibration, although the average value of the bigger budgets is lower than at the beginning, which suggests that calibration can be affected by the available budget. This tendency, however, is not present in the rest of the datasets.

\textbf{Diabetes dataset} is the largest in our evaluation, with high dimensionality and substantial class imbalance. Regarding performance, all methods converge to remarkably similar accuracy values (60-63\%) regardless of feature budget, with minimal improvement beyond 4-5 features. This flat performance landscape suggests that the most discriminative features provide limited predictive power, and additional features offer marginal information gain. We see the importance of the backbone classifier again, as CART ends up being the least performing one, which showcases the trade-off of explainability and performance. We can also see here that the uniform costs setting seems to be significantly more successful than the variable costs, which again showcases the importance of the heuristic used for each situation.
Unlike other datasets, Diabetes shows some methods maintaining remarkably stable ECE values across all budgets. VIP and RL-ReMLP hold ECE around 0.02, representing the best calibration observed across all datasets and methods. This is particularly notable as both are also the most accurate DFS methods overall, ruling out the possibility that good calibration merely reflects overly conservative base-rate estimation. This stability suggests that well-adapted methods on uniformly weak features can learn to consistently express the dataset's irreducible aleatoric uncertainty regardless of acquisition budget.

\textbf{Cirrhosis dataset.} It represents a three-class clinical problem with moderate dimensionality and limited training data. This is another case where the CART-DFS method works well for uniform feature costs but lags behind other methods in variable feature costs. We also see that the methods are not always monotonic in their performance. CWCF shows an unusually poor performance in the uniform setting, which we attribute to the RL-trained policy struggling on smaller datasets with limited training samples. Other RL approaches used a retrained backbone, which protects them from this instability in the predictor training phase. Regarding ECE, this is completely erratic in this dataset, as all methods exhibit substantial ECE fluctuations.

\textbf{Yeast dataset.} It presents the most challenging classification problem with 10 classes and minimal feature dimensionality. In this case, querying for new features is less important than the base classifier. This, for example, leaves a performance gap of 10 points between DIME and CWCF when all the features are queried. ECE here is also quite erratic.

\subsection{Prediction confidence across budgets}
\begin{figure}[ht]
    \centering
    \includegraphics[width=\linewidth]{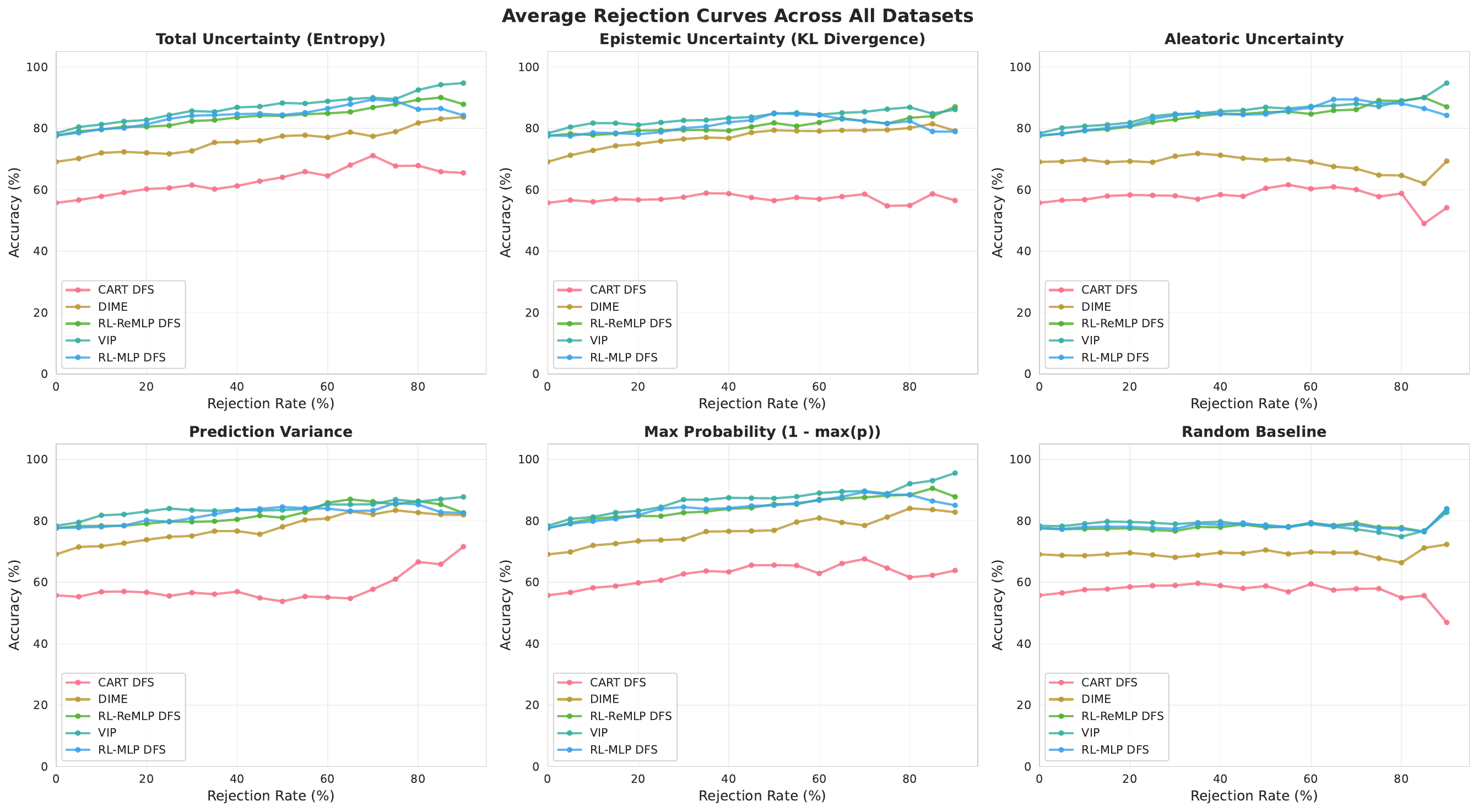}
    \caption{Evolution of accuracy rejection curves based on different uncertainty measures. Results shown for 50\% of the budget, averaged across all tabular datasets. }
    \label{fig:acc_curve_30}
\end{figure}

Here, we evaluate the quality of different uncertainty measures through accuracy-rejection curves for each uncertainty type: aleatoric (computed as in Eq. (\ref{eq:AU})), epistemic (using Eq. (\ref{eq:EU})), total uncertainty (predictive entropy), ensemble variance, 1-max probability, and a random rejection baseline. To construct an accuracy-rejection curve, we progressively reject instances with the highest uncertainty values and measure the remaining accuracy.
Well-calibrated uncertainty estimates should produce monotonically increasing curves: as we reject the most uncertain instances, accuracy on the retained set should improve. Flat or declining curves indicate poor calibration, where the uncertainty measure fails to identify genuinely difficult instances. We show these on three of our DFS proposals (Greedy CART, RL-MLP, and RL-ReMLP) and two state-of-the-art methods in the literature, DIME and VIP.  Figure \ref{fig:acc_curve_30} shows this for a 50\% budget, which was also representative of the results obtained with the rest of the budgets.

Total predictive uncertainty (entropy) and confidence-based scores (1 - max probability) produce the steepest and most consistently monotonic accuracy gains as the rejection rate increases, a pattern that remains stable across budgets and methods. In contrast, epistemic and aleatoric components have substantially flatter curves, especially for DIME and CART-DFS. Epistemic uncertainty and prediction variance behave similarly, suggesting that the auxiliary ensemble rarely induces large deviations from the primary model. Aleatoric uncertainty provides a steep curve for the best-performing methods. Overall, this is consistent with the view that neither source of uncertainty alone captures the full difficulty of a prediction under partial observation; their combination, however, provides a more robust metric for prediction reliability \citep{ovadia2019can}.

Table~\ref{tab:tabular_averaged_summary} summarises quantitative results averaged across all budgets and uncertainty-rejection curves. Lower AURC (Area Under Risk-Coverage curve) indicates better selective prediction performance, while higher AUROC measures the ability to discriminate between correct and incorrect predictions using uncertainty estimates; Risk@X\% denotes the error rate when the model rejects the (100-X)\% least confident predictions. VIP achieves the best overall uncertainty quantification performance with the lowest AURC (22.47) and highest AUROC (76.19), demonstrating superior calibration and discriminative power in identifying unreliable predictions. RL-ReMLP DFS shows competitive performance with AURC (23.81) and strong AUROC (74.59), validating its effectiveness as a DFS approach that maintains high-quality uncertainty estimates. RL-MLP DFS exhibits a notable drop in discriminative power (AUROC: 64.83) compared to RL-ReMLP DFS, with higher AURC (25.19) and elevated risk at high coverage thresholds (Risk@80\%: 30.75, Risk@90\%: 32.13), confirming the benefits of reparametrization for uncertainty quantification. DIME shows moderate AURC (27.16) with relatively strong AUROC (71.50), though risk metrics (Risk@80\%: 33.35, Risk@90\%: 34.54) indicate less reliable selective prediction at high coverage levels. CART DFS demonstrates the weakest uncertainty calibration across all metrics (AURC: 30.68, AUROC: 66.41, Risk@80\%: 34.10, Risk@90\%: 33.92), suggesting that the interpretable tree-based backbone comes at a cost to uncertainty quantification performance.

\begin{table}[htbp]
\centering
\caption{Selective Prediction Metrics for Tabular Data (Averaged Across Budgets). AURC/E-AURC/Risk: lower is better, AUROC: higher is better.}
\label{tab:tabular_averaged_summary}
\begin{tabular}{lrrrrr}
\toprule
 & AURC & E-AURC & AUROC & Risk@80\% & Risk@90\% \\ 
\midrule
VIP & 22.47 & 15.61 & 76.19 & 26.18 & 28.04 \\
RL-ReMLP DFS & 23.81 & 16.83 & 74.59 & 28.07 & 29.76\\
RL-MLP DFS & 25.19 & 17.09 & 64.83 & 30.75 & 32.13 \\
DIME & 27.16 & 19.26 & 71.50 & 33.35 & 34.54 \\
CART DFS & 30.68 & 23.08 & 66.41 & 34.10 & 33.92 \\
\bottomrule
\end{tabular}
\end{table}

\subsection{Prediction Bias Under Partial Feature Observation} \label{apx:adaptation}

\begin{figure}[ht]
    \centering

    \begin{subfigure}[t]{0.30\linewidth}
        \centering
        \includegraphics[width=\linewidth]{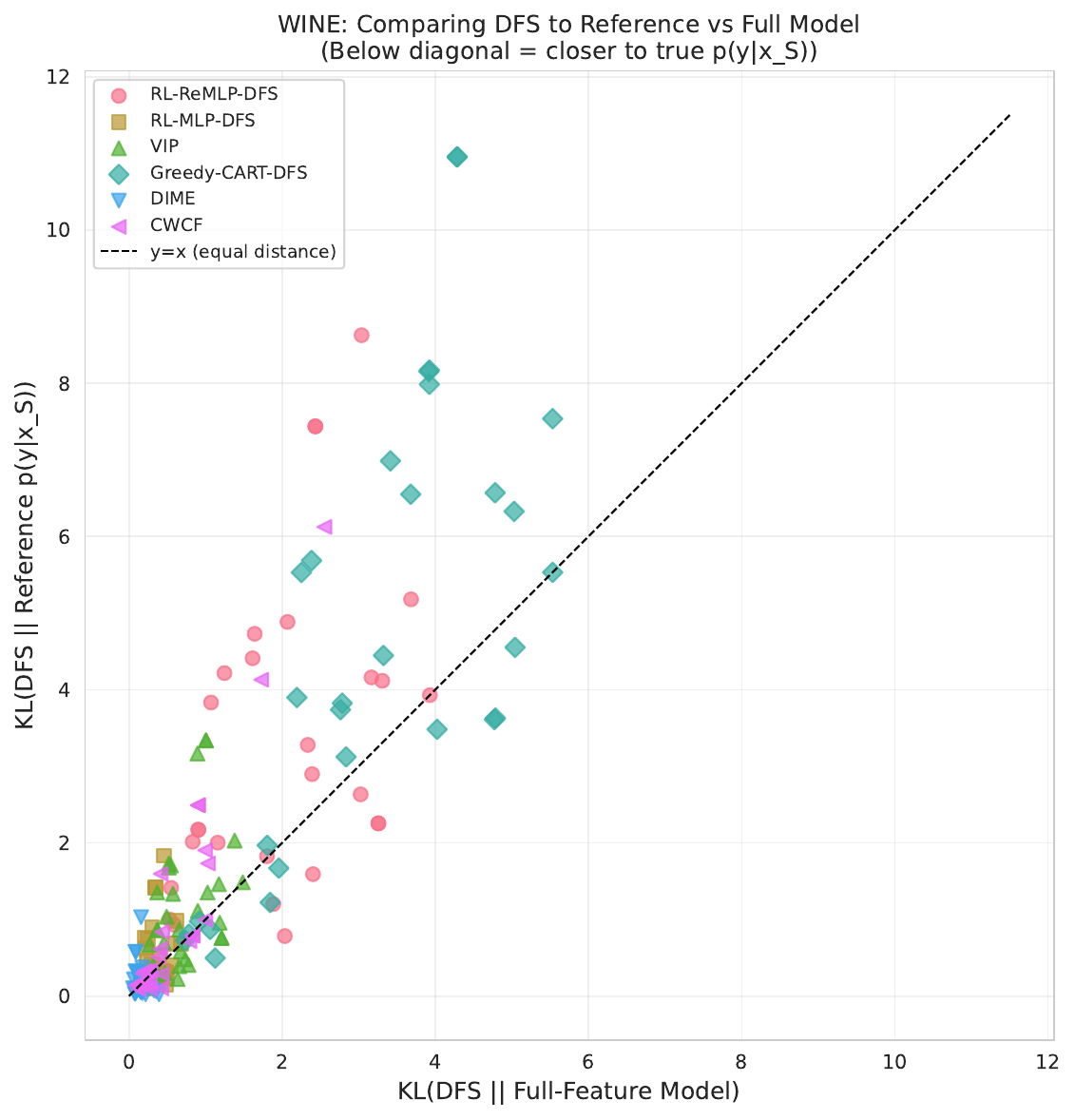}
        \caption{Wine}
        \label{fig:kl_scatter_wine}
    \end{subfigure}
    \begin{subfigure}[t]{0.30\linewidth}
        \centering
        \includegraphics[width=\linewidth]{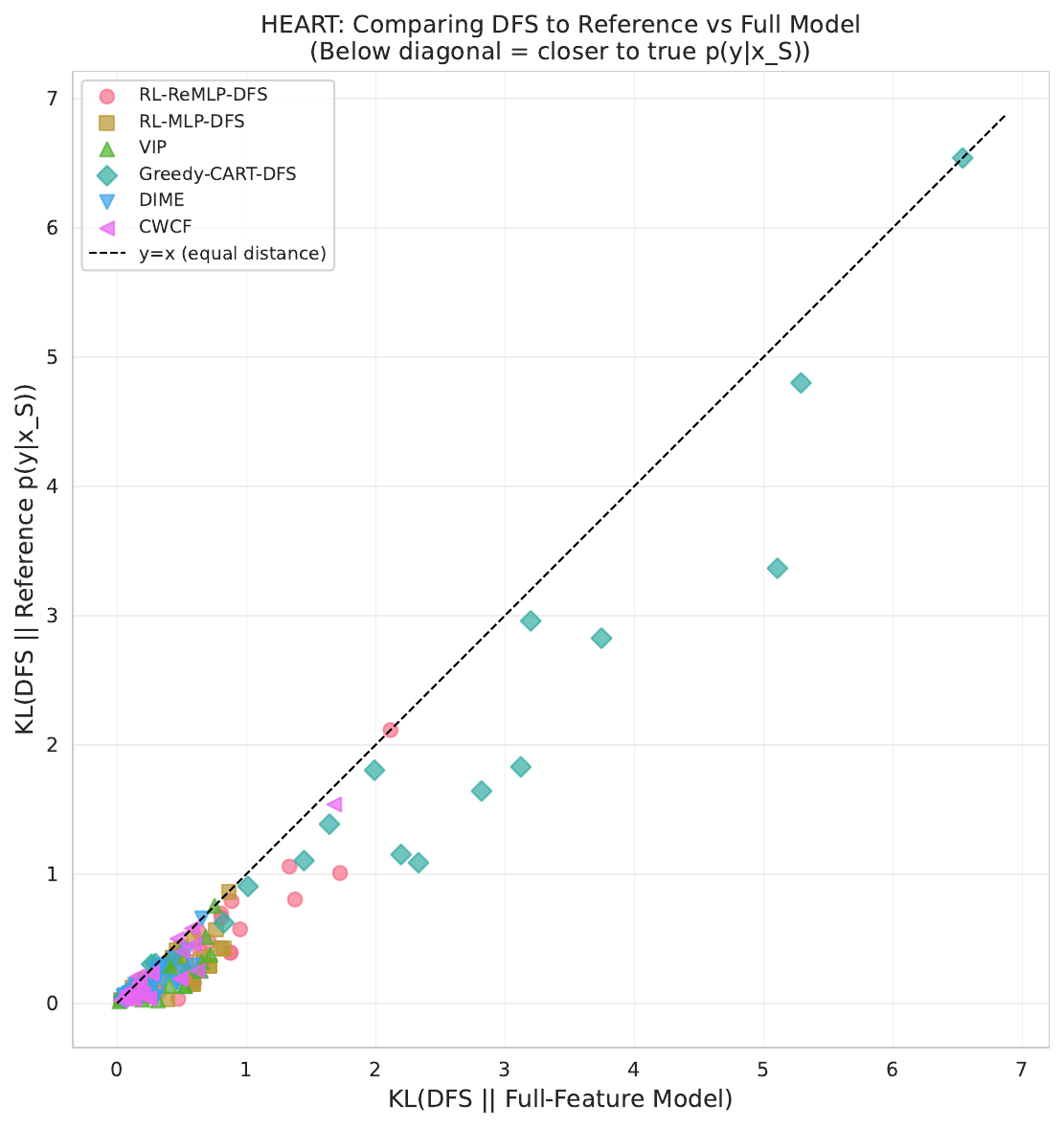}
        \caption{Heart}
        \label{fig:kl_scatter_heart}
    \end{subfigure}
    \begin{subfigure}[t]{0.30\linewidth}
        \centering
        \includegraphics[width=\linewidth]{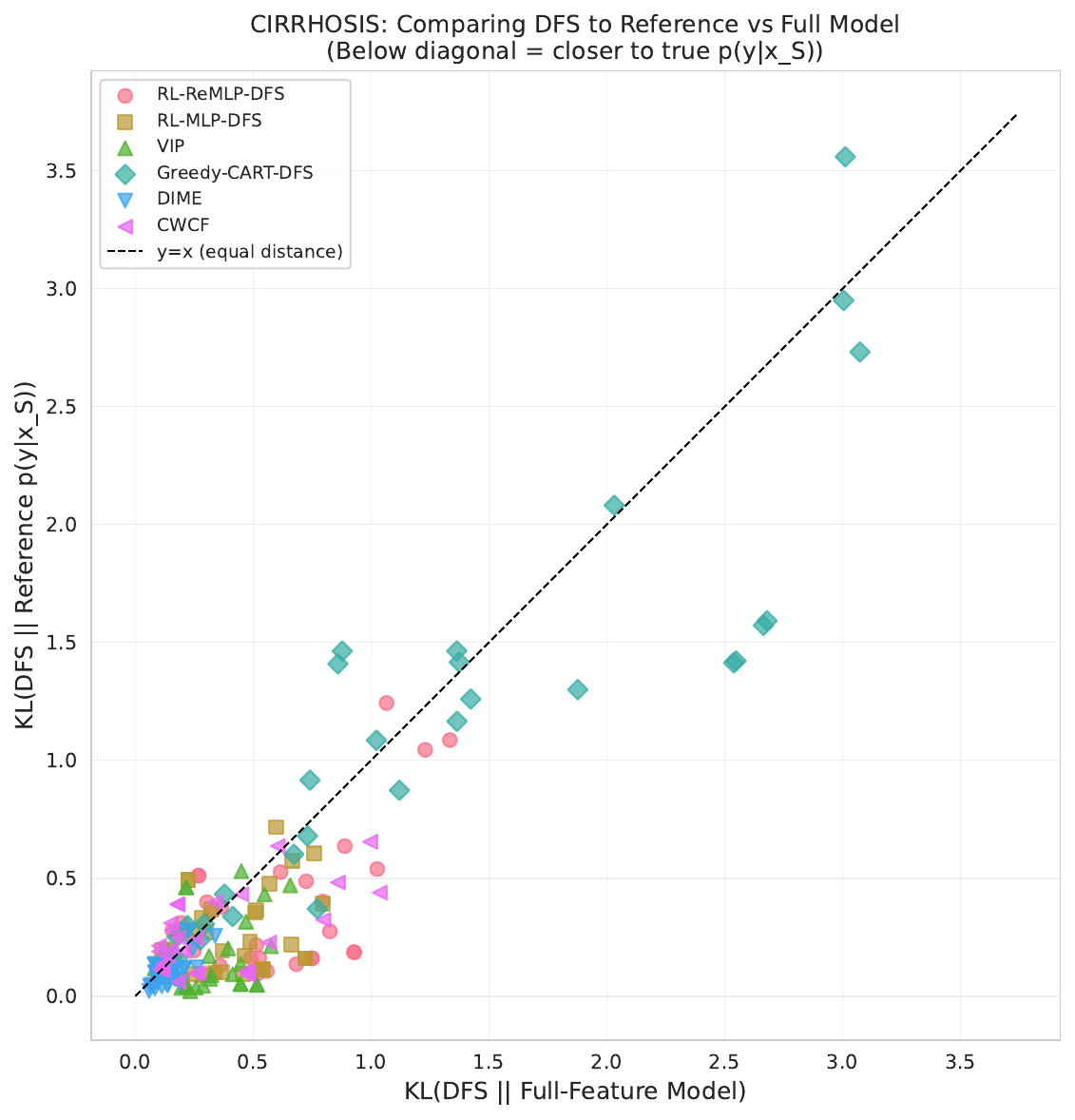}
        \caption{Cirrhosis}
        \label{fig:kl_scatter_cirrhosis}
    \end{subfigure}
    
    \begin{subfigure}[t]{0.30\linewidth}
        \centering
        \includegraphics[width=\linewidth]{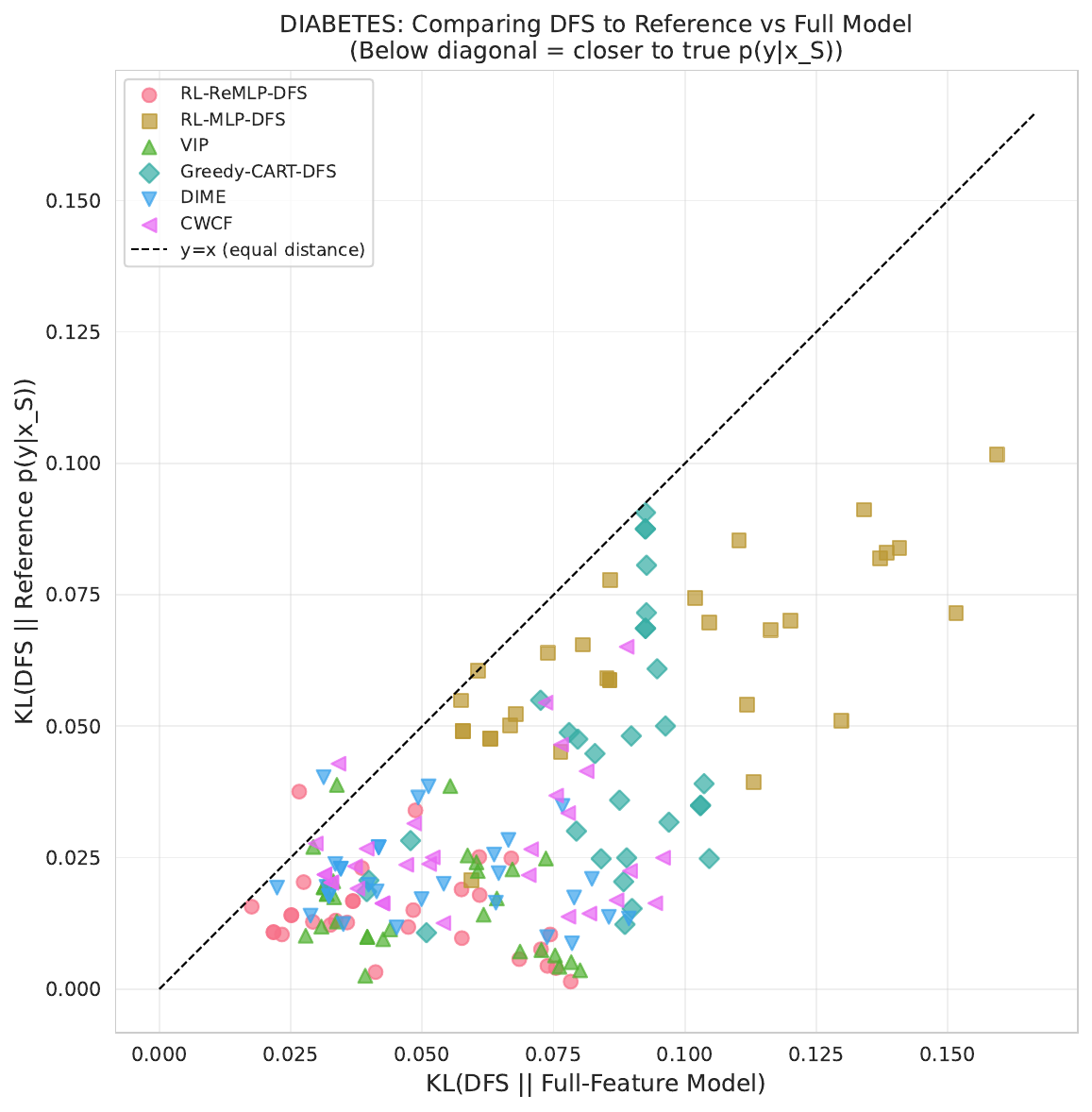}
        \caption{Diabetes}
        \label{fig:kl_scatter_diabetes}
    \end{subfigure}
    \begin{subfigure}[t]{0.30\linewidth}
        \centering
        \includegraphics[width=\linewidth]{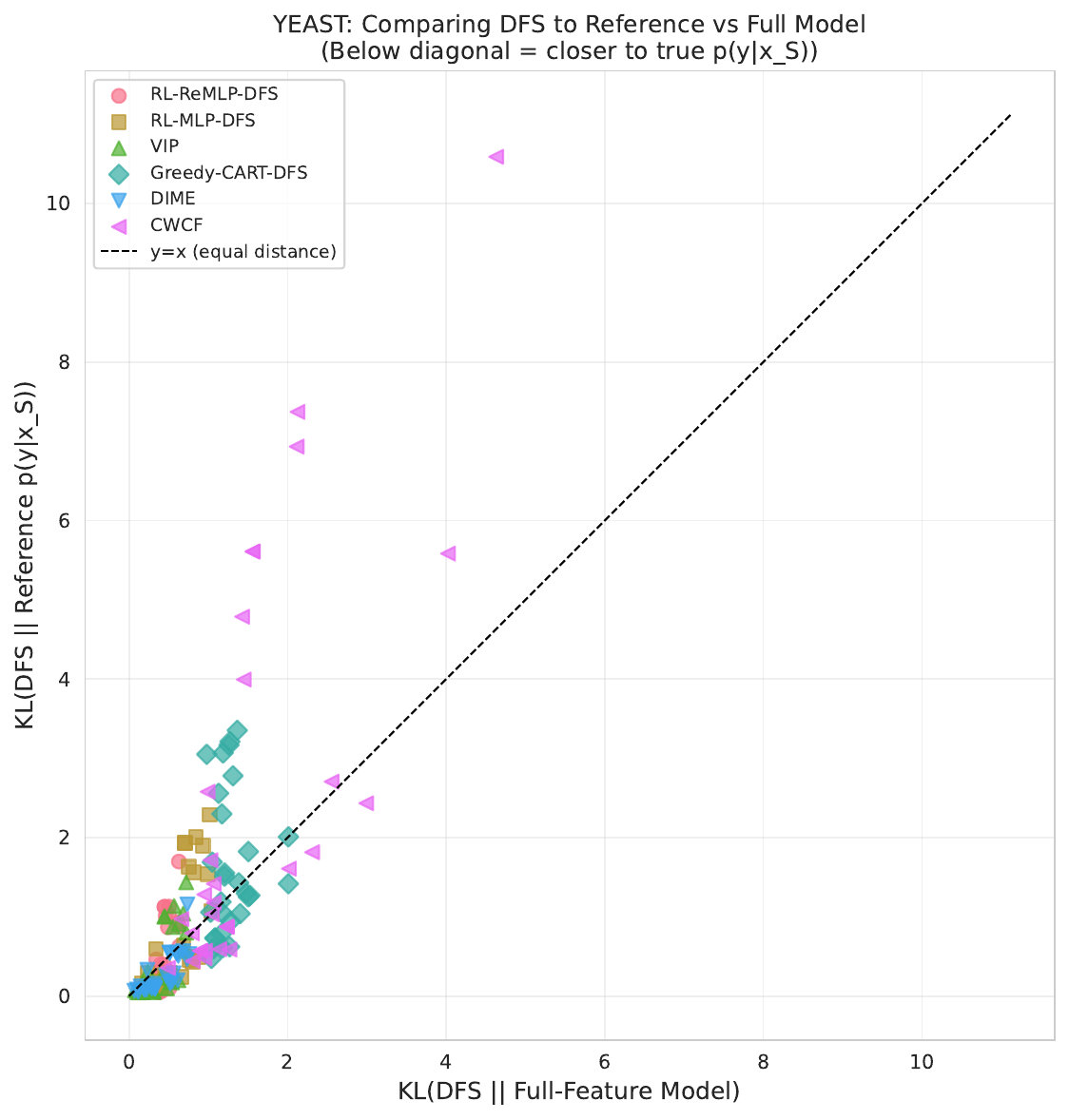}
        \caption{Yeast}
        \label{fig:kl_scatter_yeast}
    \end{subfigure}

\caption{
Bias decomposition of DFS predictions across datasets.
Each subfigure plots $D_{\mathrm{KL}}\!\left(f_{\theta^*}(\mathbf{x}_S)\,\|\,f_{\theta^*}(\mathbf{x})\right)$ (x-axis) against
$D_{\mathrm{KL}}\!\left(f_{\theta^*}(\mathbf{x}_S)\,\|\,f_{\theta_S^*}(\mathbf{x}_S)\right)$ (y-axis).
Points below the diagonal indicate good subset adaptation,
while points above indicate a stronger bias toward the full-information model.}
\label{fig:kl_scatter_all}
\end{figure}

To assess how DFS methods approximate the true conditional distribution under partial observation, we compare their predictions against two references: (i) a subset-adaptive classifier trained exclusively on $x_S$, used as a proxy for $p(y \mid x_S)$, and (ii) the full-information model trained on all features. For each method and dataset, we compute $D_\mathrm{KL}\left(f_{\theta^*}(\mathbf{x}_S) \|\ f_{\theta^*_S}(\mathbf{x}_S)\right)$ and $ D_\mathrm{KL}\left(f_{\theta^*}(\mathbf{x}_S) \|\ f_{\theta^*}(\mathbf{x})\right).$
The first term quantifies subset-adaptation error, while the second captures bias toward the full-information model.

Figure~\ref{fig:kl_scatter_all} plots these quantities against each other. Points below the diagonal indicate closer agreement with the subset-adaptive reference, whereas points near or above the diagonal suggest imperfect subset adaptation or excessive similarity to the full model. Across datasets, the relative position of methods with respect to the diagonal varies substantially, confirming that subset adaptation and approximation to the full model are distinct phenomena that depend on both dataset structure and DFS strategy. On Wine and Cirrhosis, several methods lie predominantly above the diagonal, indicating that imperfect subset adaptation is the dominant source of error. In contrast, on Diabetes, most methods cluster near the origin, suggesting that small feature subsets already recover much of the predictive signal, leaving limited room for either adaptation error or leakage. Yeast exhibits the highest dispersion, as the label structure in which both subset adaptation and approximation errors coexist and vary across methods.

Comparing the relative distance of points to the diagonal reveals systematic differences in how DFS methods handle missing information. Methods such as DIME consistently appear below or near the diagonal across datasets, indicating consistency with subset-adaptive reference predictions. In contrast, several RL-based and variational approaches exhibit points closer to the diagonal or below it in datasets where subset adaptation should induce stronger deviations, suggesting partial reconstruction of missing information through learned correlations. Importantly, large deviations from both references, as observed for Greedy CART-DFS on Wine and Yeast, are not indicative of leakage but rather of unstable subset adaptation, emphasising that low agreement with the full model alone is insufficient to diagnose correct DFS behaviour. Overall, these results reinforce the need to jointly evaluate divergence to both subset-adaptive and full-information references when assessing DFS methods, as reliance on either comparison in isolation can lead to misleading conclusions about adaptation quality and information usage.



\section{Targeted Validation Experiments.}
\label{sec:appendix_strengthening}

We provide four additional experiments to further validate robustness under distribution shift, the relationship between epistemic uncertainty and subset adaptation risk, the effect of last-layer reparametrization on reducing adaptation error and the correlation between aleatoric and epistemic uncertainty. For these tests, we consider our RL-based DFS policy with the following backbones: CART, MLP, ReMLP, AdMLP, AdpReMLP. We use two budget levels: up to 30\% and up to 70\% of the total features.

\subsection{Robustness Under Distribution Shift}
\label{sec:appendix_shift}

Here, we test if epistemic uncertainty increases under different distributional perturbations and how much predictive performance and epistemic uncertainty are affected by them. Perturbations are applied only to the test set to simulate distribution shift: for 50\% of features, additive Gaussian noise $\varepsilon \sim \mathcal{N}(0,\sigma^2)$ is applied with $\sigma \in \{0.1, 0.2, 0.8, 2.0\}$. For each model and budget level, we compute accuracy, ECE and mean predictive entropy. A consistent increase in entropy and selective prediction performance under perturbation indicates that uncertainty estimates respond appropriately to distribution shift.

Table~\ref{tab:shift_results} reports the averaged results across datasets. Under mild covariate noise ($\sigma = 0.1$), all methods exhibit negligible accuracy loss (below 3 points on average), confirming that DFS predictions are robust to small input perturbations. At $\sigma = 0.2$, the drops remain moderate, up to 3.7 points for CART-DFS at the 70\% budget, and no method degrades catastrophically. A stronger perturbation, $\sigma = 0.8$, reveals clearer separation: CART-DFS drops to $.61$ and $.66$ at the 30\% and 70\% budgets respectively, while neural methods' performance remains less affected. Under the largest perturbation ($\sigma = 2.0$), all methods experience significant accuracy degradation. The sharpest drops occur in CART-DFS and RL-ReMLP. Methods whose architectures are jointly trained to handle partial observations are more resilient here. The epistemic uncertainty is considerably higher than in previous $\sigma$ values as well, further confirming that epistemic uncertainty reliably grows with severe distributional mismatch.

For the methods that provide epistemic variance through auxiliary models, even the mildest perturbation ($\sigma = 0.1$) produces a measurable increase in $\bar\varepsilon$. At $\sigma = 0.8$ CART-DFS epistemic variance rises by $20\%$ (from $.044$ to $.053$) at the 30\% budget and by $31\%$ (from $.045$ to $.059$) at 70\%, while RL-ReMLP increases by $50\%$ and $54\%$ respectively. Under the most severe perturbation ($\sigma = 2.0$), the epistemic response is even more pronounced: RL-ReMLP DFS variance at 70\% reaches $.038$, nearly triple its unperturbed value, while CART-DFS rises to $.068$ ($+51\%$). ECE follows a similar upward trend across all noise levels, confirming progressive calibration degradation under shift. 

These results support two conclusions. First, the DFS predictions are robust to the levels of covariate noise that might arise from measurement error, with accuracy losses below 4 points up to $\sigma = 0.2$. Second, and more importantly for the uncertainty framework, the epistemic estimates increase monotonically with perturbation severity, confirming that they capture genuine distributional mismatch rather than being mere artefacts of the ensemble.

   \begin{table}[t]                                                                                                                                                                            
  \centering
  \caption{Performance under distribution shift, averaged over all tabular
    datasets. $B$: budget level (\% of features); $e$: mean
    epistemic variance (only for methods with auxiliary models).}
  \label{tab:shift_results}
  \resizebox{\textwidth}{!}{%
  \begin{tabular}{ll ccc ccc ccc ccc ccc}
  \toprule
  & & \multicolumn{3}{c}{No shift}
  & \multicolumn{3}{c}{$\sigma{=}0.1$}
  & \multicolumn{3}{c}{$\sigma{=}0.2$}
  & \multicolumn{3}{c}{$\sigma{=}0.8$}
  & \multicolumn{3}{c}{$\sigma{=}2.0$} \\
  \cmidrule(lr){3-5}\cmidrule(lr){6-8}\cmidrule(lr){9-11}\cmidrule(lr){12-14}\cmidrule(lr){15-17}
  Method & $B$
    & Acc & ECE & $e$
    & Acc & ECE & $e$
    & Acc & ECE & $e$
    & Acc & ECE & $e$
    & Acc & ECE & $e$ \\
  \midrule
  \multirow{2}{*}{Greedy CART-DFS}
    & 30\% & .72 & .09 & .044  & .70 & .09 & .046  & .70 & .10 & .047  & .61 & .15 & .053  & .55 & .19 & .057 \\
    & 70\% & .77 & .09 & .045  & .77 & .08 & .047  & .74 & .11 & .051  & .66 & .17 & .059  & .57 & .25 & .068 \\
  \addlinespace
  \multirow{2}{*}{RL-ReMLP DFS}
    & 30\% & .74 & .07 & .004  & .73 & .08 & .004  & .73 & .09 & .005  & .66 & .12 & .006  & .59 & .22 & .013 \\
    & 70\% & .78 & .08 & .013  & .78 & .09 & .014  & .76 & .09 & .015  & .70 & .18 & .020  & .60 & .32 & .038 \\
  \addlinespace
  \multirow{2}{*}{DIME}
    & 30\% & .72 & .08 & ---   & .70 & .08 & ---   & .70 & .08 & ---   & .66 & .15 & ---   & .61 & .21 & --- \\
    & 70\% & .78 & .08 & ---   & .78 & .07 & ---   & .77 & .08 & ---   & .74 & .10 & ---   & .64 & .22 & --- \\
  \addlinespace
  \multirow{2}{*}{VIP}
    & 30\% & .68 & .23 & ---   & .69 & .23 & ---   & .68 & .22 & ---   & .68 & .21 & ---   & .61 & .12 & --- \\
    & 70\% & .75 & .24 & ---   & .75 & .24 & ---   & .75 & .23 & ---   & .73 & .21 & ---   & .68 & .17 & --- \\
  \bottomrule
  \end{tabular}%
  }
  \end{table}

\subsection{Epistemic Uncertainty and Subset Risk Gap}
\label{sec:appendix_deltaS}

Here, we empirically assess whether epistemic uncertainty correlates with the subset adaptation risk gap  $\Delta_S = R_S\!\left(f_{\theta^*}\right) - R_S\!\left(f_{\theta_S^*}\right)$
introduced in Section~\ref{sec:uncertainty_in_dfs}. For a subset $S$ with $|S| = b$, we denote 
$R_S\!\left(f_{\theta^*}\right)$ as the test error of the globally trained model evaluated under mask $S$, 
$R_S\!\left(f_{\theta_S^*}\right)$ as the test error of a model trained exclusively on features $S$, 
and $\bar{e}_S$ as the mean epistemic uncertainty across test samples under mask $S$. 
Then, for each budget level, we sample $K=100$ random subsets of size $b$. 
For each subset, we compute $(\Delta_S, \bar{e}_S)$ and report Pearson and Spearman correlations across all sampled subsets. 
A positive correlation indicates that epistemic uncertainty reliably identifies subsets for which global masking induces substantial performance degradation relative to a locally adapted oracle.

\begin{figure}
    \centering
    \includegraphics[width=0.85\linewidth]{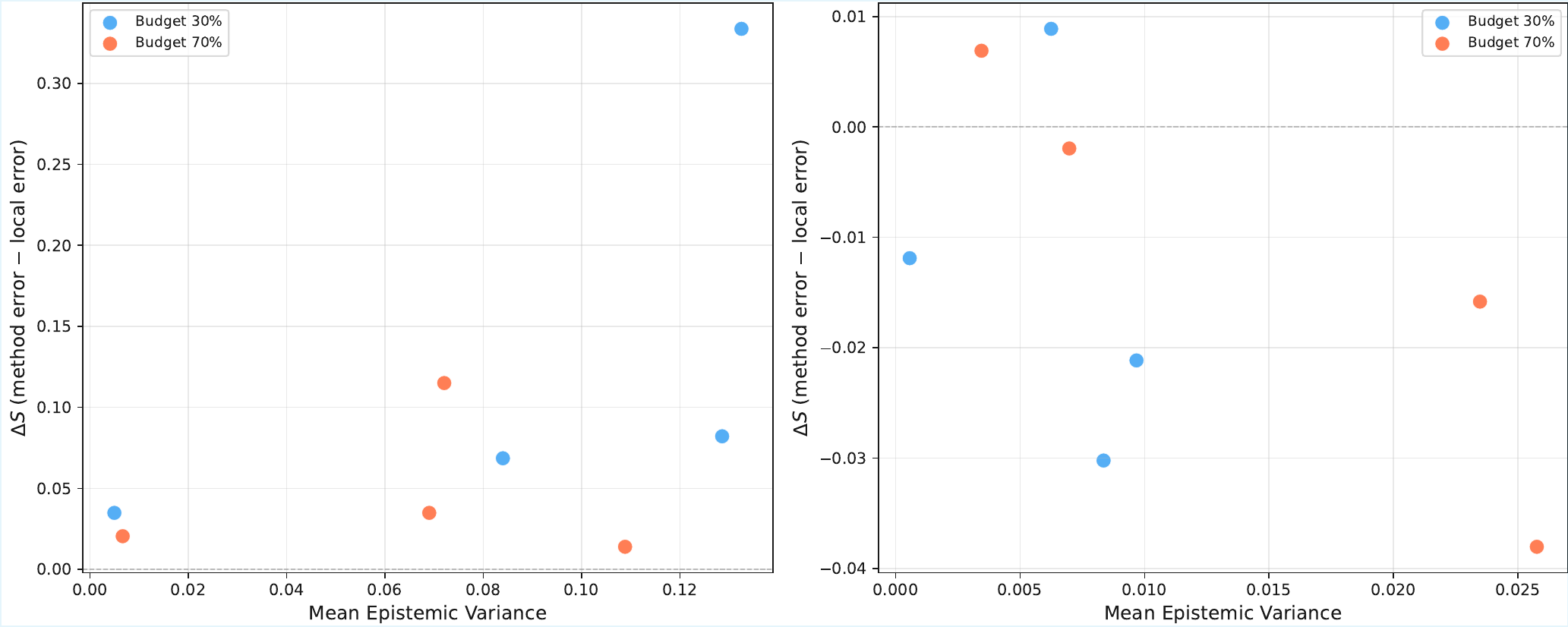}
    \caption{Scatter plot of $\Delta_S$ versus mean epistemic variance
    $\bar{e}_S$ on average of all datasets ($K{=}100$ subsets per budget).
    Left: CART-DFS shows a strong positive correlation
    ($r{=}0.86$, $p{<}0.001$): subsets where the global model incurs a larger gap over the local oracle are those affected by higher epistemic uncertainty.
    Right: RL-ReMLP DFS exhibits a weak correlation ($r{=}0.25$),
    consistent with its near-zero $\Delta_S$; the reparametrized model
    leaves little adaptation gap for the uncertainty signal to track.}
  \label{fig:deltaS_scatter}

\end{figure}

Table~\ref{tab:deltaS_corr} and Figure~\ref{fig:deltaS_scatter} report the results. 
For Greedy CART-DFS, the correlation between $\Delta_S$ and $\bar{e}_S$ is strongly positive on three of four datasets 
(wine: $r{=}0.86$; cirrhosis: $r{=}0.51$; heart: $r{=}0.34$; all $p{<}0.001$), 
with an average Pearson $r$ of $0.39$ and Spearman $\rho$ of $0.39$. 
For RL-ReMLP-DFS, correlations are weaker (average $r{=}0.15$), with mixed signs across datasets. 
This is explained by the method's characteristically small $\Delta_S$: averaged across datasets, 
$\bar{\Delta}_S = -0.01$ at both budget levels, indicating that the reparametrized global model matches or slightly outperforms locally trained oracles. 
When the adaptation gap is near zero, there is little signal for any uncertainty proxy to track.

  \begin{table}[t]
  \centering
  \caption{Pearson ($r$) and Spearman ($\rho$) correlations between
    $\Delta_S$ and mean epistemic variance $\bar{e}_S$ across $K{=}100$
    random subsets per budget level. All per-dataset correlations for
    CART\,DFS are significant at $p < 0.05$ except yeast ($p{=}0.034$,
    negative). All RL-ReMLP correlations are significant at $p < 0.01$.}
  \label{tab:deltaS_corr}
  \begin{tabular}{l cc cc}
  \toprule
  & \multicolumn{2}{c}{CART DFS}
  & \multicolumn{2}{c}{RL-ReMLP DFS} \\
  \cmidrule(lr){2-3}\cmidrule(lr){4-5}
  Dataset & $r$ & $\rho$ & $r$ & $\rho$ \\
  \midrule
  Wine      &  .86 &  .86 &  .25 &  .27 \\
  Heart     &  .34 &  .34 & -.26 & -.23 \\
  Cirrhosis &  .51 &  .55 &  .20 &  .23 \\
  Yeast     & -.15 & -.18 &  .43 &  .40 \\
  
  \addlinespace
  Average   &  .39 &  .39 &  .15 &  .17 \\
  \bottomrule
  \end{tabular}
  \end{table}

\subsection{Effect of Reparametrization on Subset Risk}
\label{sec:appendix_reparam}

In this experiment, we test how much last-layer reparametrization reduces the subset risk gap $\Delta_S$. For each sampled subset $S$, we compute:
\begin{align}
\Delta_S^{\text{MLP}} 
&= R_S\!\left(f_{\theta^*}^{\text{MLP}}\right) 
   - R_S\!\left(f_{\theta_S^*}^{\text{MLP}}\right), \\
\Delta_S^{\text{ReMLP}} 
&= R_S\!\left(f_{\theta^*}^{\text{ReMLP}}\right) 
   - R_S\!\left(f_{\theta_S^*}^{\text{ReMLP}}\right).
\end{align}
We compare the distributions of $\Delta_S^{\text{MLP}}$ and $\Delta_S^{\text{ReMLP}}$ across all subsets and budgets using the Wilcoxon signed-rank test. We additionally report mean and median risk gaps and the fraction of subsets for which reparametrization reduces $|\Delta_S|$. A systematic reduction of $\Delta_S$ confirms that the adapter mitigates the bias induced by mean-imputation under arbitrary subsets.

\begin{figure}[t]
    \centering
    
    \begin{subfigure}[t]{0.30\linewidth}
        \centering
        \includegraphics[width=\linewidth]{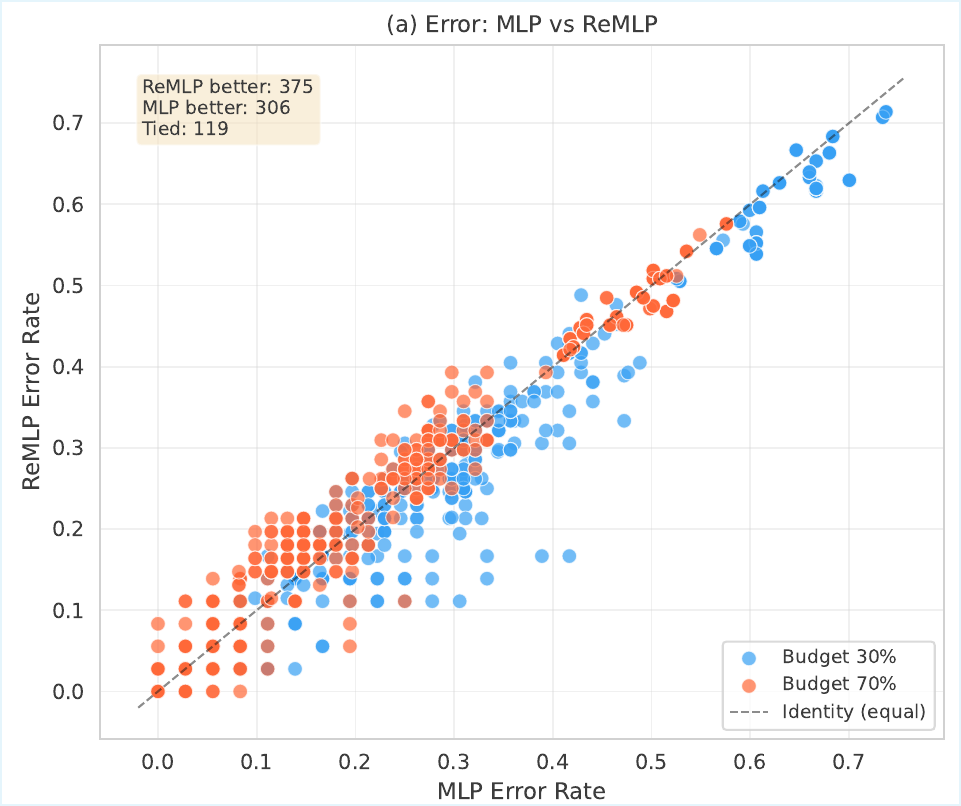}
        \caption{}
        \label{fig:delta_a}
    \end{subfigure}
    \hfill
    \begin{subfigure}[t]{0.30\linewidth}
        \centering
        \includegraphics[width=\linewidth]{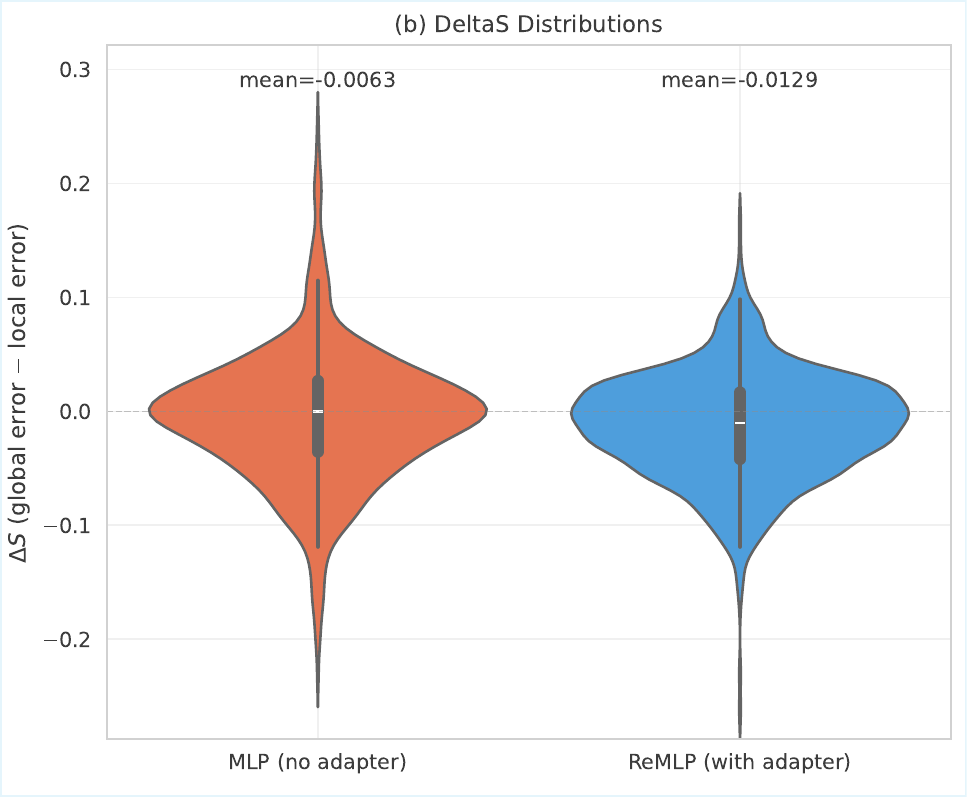}
        \caption{}
        \label{fig:delta_b}
    \end{subfigure}
    \hfill
    \begin{subfigure}[t]{0.30\linewidth}
        \centering
        \includegraphics[width=\linewidth]{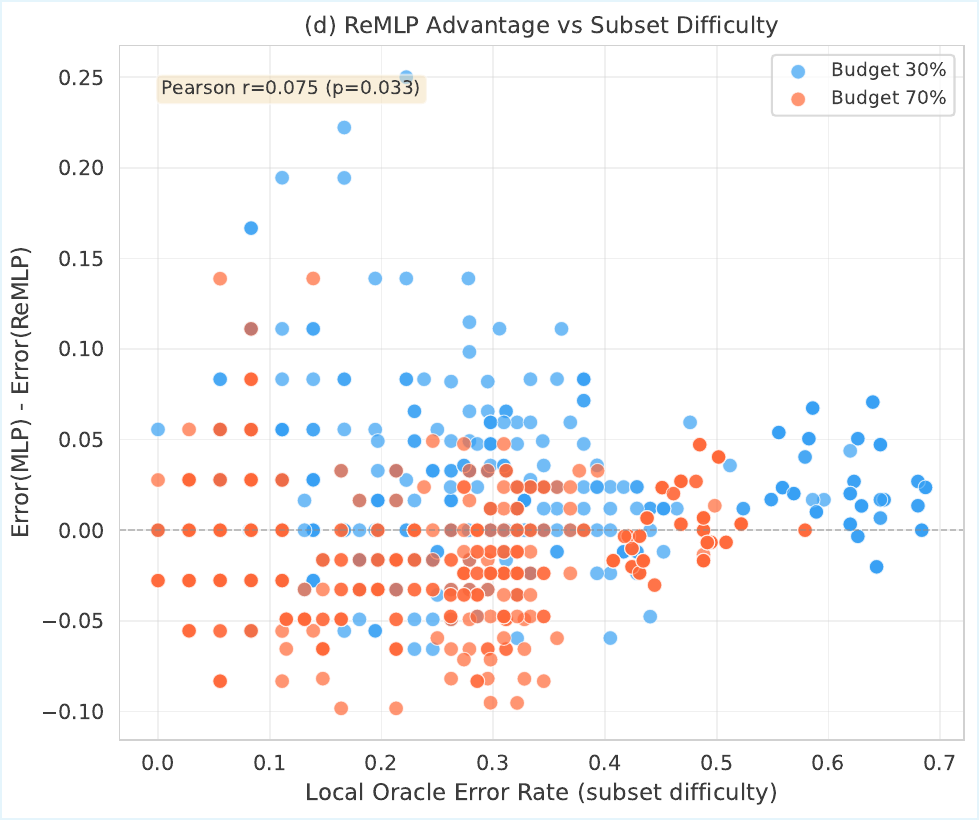}
        \caption{}
        \label{fig:delta_c}
    \end{subfigure}
  \caption{Reparametrization analysis across 800 sampled subsets
  (4 datasets $\times$ 2 budgets $\times$ 100 subsets).
  \textbf{(a)}~Scatter of per-subset error rates: points below the identity line indicate ReMLP outperforms MLP.
  \textbf{(b)}~Violin plots of $\Delta_S$ for each model; the
  ReMLP distribution is narrower and shifted further below zero.
  \textbf{(c)}~ReMLP advantage as a function of subset difficulty
  (local oracle error).}
  \label{fig:reparam}
  \end{figure}

Table~\ref{tab:reparam_deltaS} reports the signed and absolute risk gaps for each dataset. Across all 800 sampled subsets, the reparametrized model (ReMLP) achieves a mean error rate 0.66 percentage points lower than the standard MLP (Wilcoxon signed-rank $p < 0.001$; Figure~\ref{fig:reparam}c), outperforming it on 47\% of subsets versus 38\% for the MLP (15\% tied).

Comparing the absolute risk gaps, $|\Delta_S^{\text{ReMLP}}|$ is significantly smaller than $|\Delta_S^{\text{MLP}}|$ overall (0.037 vs.\ 0.041, Wilcoxon $p = 0.012$), confirming that the adapter narrows the gap between global and locally trained models. The effect is most pronounced on \emph{wine}, where the adapter nearly eliminates the positive gap at 30\% budget ($\Delta_S^{\text{MLP}} = +0.056 \to \Delta_S^{\text{ReMLP}} = +0.009$; Wilcoxon $p = 0.008$), and is also significant on \emph{heart} ($p = 0.048$). On \emph{cirrhosis} and \emph{yeast}, the reduction is not statistically significant, which we attribute to both models already achieving near-local performance on these datasets (mean $|\Delta_S| < 0.02$ on yeast).

The violin plot (Figure~\ref{fig:reparam}b) further illustrates the narrower $\Delta_S$ distribution of the reparametrized model: while the MLP distribution is centered near zero ($\overline{\Delta_S} = -0.006$), the ReMLP distribution is shifted further below zero ($\overline{\Delta_S} = -0.013$), indicating that the adapter not only reduces the gap but systematically improves upon local specialists. This could indicate that global behaviour plus local reparametrization can achieve a regularisation effect, or that some leakage with respect to the full prediction could be happening.

  \begin{table}[t]
  \centering
  \caption{Reparametrization effect on the subset risk gap.
  $\Delta_S$ is the signed gap (positive = global worse than local);
  $|\Delta_S|$ is the absolute gap.  ``Wins'' reports the fraction
  of subsets where $|\Delta_S^{\text{ReMLP}}| <
  |\Delta_S^{\text{MLP}}|$.  Wilcoxon $p$-values test whether
  $|\Delta_S|$ differs between the two models.}
  \label{tab:reparam_deltaS}
  \small
  \begin{tabular}{lcccccc}
  \toprule
   & \multicolumn{2}{c}{Mean $\Delta_S$}
   & \multicolumn{2}{c}{Mean $|\Delta_S|$}
   & Wins & Wilcoxon \\
  \cmidrule(lr){2-3} \cmidrule(lr){4-5}
  Dataset & MLP & ReMLP & MLP & ReMLP & (\%) & $p$ \\
  \midrule
  Wine & $+.028$ & $+.004$ & .051 & .040 & 39.5 & .008 \\
  Heart& $-.040$ & $-.030$ & .053 & .047 & 47.0 & .048 \\
  Cirrhosis & $-.025$ & $-.023$ & .041 & .042 & 40.0 & .706 \\
  Yeast& $+.012$ & $-.003$ & .019 & .018 & 39.5 & .734 \\
  \midrule
  \textbf{All} & $-.006$ & $-.013$ & .041 & .037 & 41.5 & .012 \\
  \bottomrule
  \end{tabular}
  \end{table}

\subsection{Relationship between Epistemic and Aleatoric Uncertainty}

\begin{figure}[ht]
    \centering
    \includegraphics[width=0.90\linewidth]{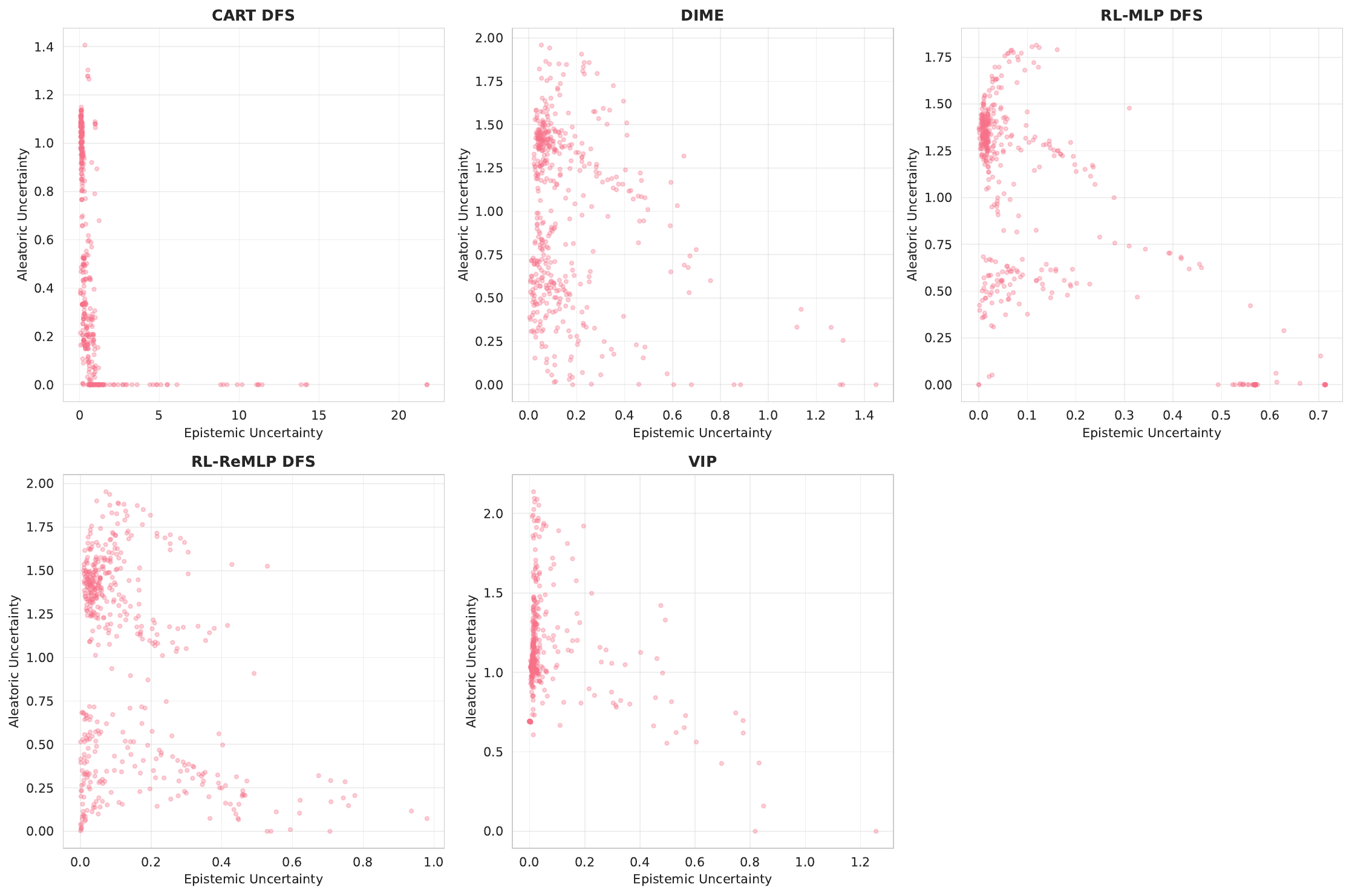}
    \caption{Relationship between aleatoric and epistemic uncertainty, averaged across all datasets and budgets, for the different methods shown.}
    \label{fig:random_epistemic_relationship}
\end{figure}

We examine how aleatoric and epistemic uncertainty jointly
distribute across instances, computed according to
Eqs.~(\ref{eq:AU}) and~(\ref{eq:epistemic_normal}). Figure~\ref{fig:random_epistemic_relationship}
displays scatter plots of both quantities across all experiments and DFS
methods. For most models, epistemic uncertainty remains predominantly low,
while aleatoric uncertainty is more broadly spread across instances. Notably,
instances with the highest aleatoric uncertainty tend to exhibit low epistemic
uncertainty, and vice versa. This result differs from findings in prior
work \citep{valdenegro2022deeper}, but is not entirely surprising. In
high aleatoric cases, significant class overlap causes auxiliary models to
quickly converge to conservative and similar predictions. Conversely, when epistemic uncertainty is high, aleatoric uncertainty appears difficult to estimate properly. This mutual exclusivity is particularly striking in VIP, where instances with jointly high or jointly low values of both uncertainties are nearly absent.

\section{Implementation Details and Reproducibility}
\label{appendix:report}

\subsection{Dataset Statistics} \label{appendix:report_datasets}

\begin{table}[ht]
\centering
\caption{Dataset Summary}
\label{tab:datasets}
\begin{tabular}{lrrrr}
\toprule
\textbf{Dataset} & \textbf{Samples} & \textbf{Features} & \textbf{Classes} &\textbf{Acc. LR}  \\
\midrule
Diabetes & 101,766 & 47 & 2 & 0.61\\
Heart & 303 & 13 & 2 & 0.86 \\
Cirrhosis & 520 & 17 & 3 & 0.72 \\
Wine & 178 & 13 & 3 & 0.98 \\
Yeast & 1,484 & 8 & 10 & 0.57\\
\bottomrule
\end{tabular}
\end{table}

A summary of the classification datasets' size and properties is in Table \ref{tab:datasets}. We also included the performance of a logistic regressor for each one as a baseline reference. 

\begin{itemize}
    \item \emph{Diabetes} dataset \citep{strack2014impact}: has 101766 samples and 47 features regarding patient condition. The target value is if the patient is readmitted to the hospital in a $30$ day period after the last discharge. 
    \item \emph{Heart} disease \citep{Detrano1989InternationalAO}: contains 13 clinical features for each of the 303 patients. The target field refers to the presence of heart disease in the patient, which has been binarised into presence and absence. 
    \item \emph{Cirrhosis} Patient Survival Prediction \citep{fleming2013counting}: uses 17 clinical features for predicting the survival state of 520 patients with liver cirrhosis. These include death, alive, or alive due to liver transplantation. 
    \item \emph{Wine}: the classical dataset containing chemical analysis results of wines \citep{wine_109}. It includes 178 samples with 13 features, used for classification into wine types.
    \item \emph{Yeast}: protein localization sites in yeast cells \citep{yeast_110}. It consists of 1,484 samples with 8 features, predicting the cellular compartment where a protein resides.
\end{itemize}

\subsection{Value Function Estimator Architecture} \label{app:neural_estimator}

At inference time, the value function $v_q(i, \mathbf{x}_S)$ defined in Eq.~(\ref{eq:main_value_function}) cannot be directly computed because the unqueried features $x_i$ are not observed. To address this, we train a neural network estimator $h_{\phi}$ that takes the observed feature subset $\mathbf{x}_S$ as input and predicts the expected value of querying each unobserved feature.

\textbf{Architecture.} We use a single neural network with two output heads:
\begin{itemize}
    \item $h_{u,i}(\mathbf{x}_S)$: predicts the reduction in prediction discrepancy $u(\mathbf{x}_S \cup \{x_i\})$
    \item $h_{e,i}(\mathbf{x}_S)$: predicts the reduction in epistemic uncertainty $e(\mathbf{x}_S \cup \{x_i\})$
\end{itemize}

Each head has $M$ outputs (one per feature), allowing the network to predict the value of adding any unobserved feature given the current observation $\mathbf{x}_S$.

\textbf{Network Architecture Details.} For tabular experiments, we use:
\begin{itemize}
    \item Shared encoder: 2-layer MLP with hidden size 64.
    \item Activation: SELU with dropout rate 0.5.
    \item Two separate output heads, each producing $M$ values.
    \item Training: 100 epochs (comparison experiments).
    \item Learning rate: 0.01, Adam optimiser.
    \item L1 regularisation: tested $\lambda \in \{0.0, 0.001, 0.01, 0.1, 0.5, 1.0, 2.0\}$
\end{itemize}

 \subsection{Hyperparameters} \label{appendix:report_hyperparam}     

 \paragraph{Tabular Experiments.}   Our neural estimator proposed in Section 5.4 uses the following parameters: 
  \begin{itemize}             
      \item \textbf{Architecture}: 2-layer MLP with hidden size 64            
      \item \textbf{Activation}: SELU with dropout rate 0.5   
      \item \textbf{Training}: 100 epochs (comparison), 10 epochs (individual), learning rate 0.01, Adam optimiser            
      \item \textbf{Regularisation}: L1 weights tested [0.0, 0.001, 0.01, 0.1, 0.5, 1.0, 2.0] 
  \end{itemize}               

The hyperparameters for the backbone classifiers for our DFS proposal are:  
  \begin{itemize}             
      \item \textbf{Decision Tree}: minimum of 5 samples for a split, a minimum of 2 samples to generate a leaf, no maximum depth constraint  
      \item \textbf{Random Forest}: 100 estimators, same split parameters as Decision Tree    
      \item \textbf{MLP}: one hidden layer of size 100 with ReLU activation, Adam optimiser with 200 iterations     
    \end{itemize}
    
\paragraph{Reinforcement Learning Policy.}
RL-based feature selection is trained via Q-learning \citep{clifton2020q}. The state consists of the binary mask of observed features concatenated with their current values. The Q-network is a 3-layer MLP (hidden size 128, ReLU activations) with a duelling architecture, outputting Q-values for all candidate feature queries. Training uses $\epsilon$-greedy exploration ($\epsilon$ decayed from 1.0 to 0.01), discount factor $\gamma=0.99$, replay buffer size 10{,}000, batch size 32, target network updates every 100 steps, and Adam optimizer (learning rate $10^{-4}$). Rewards are defined as $R_t = q(x_i, \mathbf{x}_{S_t})$ (Eq.~\ref{eq:gain}), where $x_i$ is the observed value of the queried feature, and policies are trained for 1{,}000 episodes per dataset.
  
  \paragraph{Deep Learning Models.}
All deep learning methods are trained for 100 total epochs to ensure comparability, 
with method-specific stage allocation: VIP (100 joint), DIME (25 pretraining + 75 joint), 
Greedy-MLP (100 policy), Greedy-ReMLP (60 adapter + 40 policy), 
Adp-MLP (40 pretraining + 60 policy), and Adp-ReMLP 
(40 pretraining + 40 adapter + 20 policy).

Training uses a batch size of 64 and the Adam optimiser. Learning rates are $10^{-3}$ 
for pretraining/adapter phases and $10^{-4}$ for policy networks. 
We use random seed 42 and 30 MC-Dropout samples for uncertainty estimation. 
Images are divided into a $4\times4$ grid (16 patches) with a budget range of 2--10 patches.

All methods use a ResNet-18 backbone, pretrained on ImageNet for RGB datasets (CIFAR-10, ImageNette) and trained from scratch for MNIST. The dropout rate is set to 0.2 for MC-Dropout inference.

  \paragraph{Method-Specific Architectures.}  

  \begin{itemize}           
  \item \textbf{VIP Query Network}: Fully connected layers with hidden sizes 1000 and 500, dropout rate 0.3. 
  \item \textbf{DIME Selector Network}: Fully connected layers with hidden sizes 512 and 256, dropout rate 0.3.
  \item \textbf{Greedy Policy Estimator}: Hidden dimension 256, dropout rate 0.3, L1 regularization $\lambda = 0.1$.        
  \item \textbf{Subset Adapter (ReMLP methods)}: hidden dimension 64, generating scale and shift parameters for the final classification layer.         
  \end{itemize}           

  For Adp-MLP and Adp-ReMLP, we pretrain the classifier with random masking where 20--100\% of patches are visible per sample. This trains the model to be robust to partial observations before learning the selection policy.

\subsection{Code Availability}
All code, including custom implementations of dynamic feature selection methods, neural estimators, and evaluation frameworks, will be made publicly available upon publication.

\end{document}